\documentclass[11pt]{article}
\usepackage[preprint]{acl}
\usepackage{times}
\usepackage{latexsym}
\usepackage[T1]{fontenc}
\usepackage[utf8]{inputenc}
\usepackage{microtype}
\usepackage{inconsolata}
\usepackage{graphicx}
\usepackage{booktabs}
\usepackage{multirow}
\usepackage{makecell}
\usepackage{pifont}
\usepackage{enumitem}
\usepackage{amsmath}
\usepackage{longtable}
\usepackage[most]{tcolorbox}
\usepackage[table]{xcolor}
\usepackage{amssymb}
\usepackage{tabularx,array}
\usepackage{dsfont}

\newcommand{\belowrule}[1]{\cellcolor{black!8}#1}

\newtcolorbox{promptbox}[1][]{
    enhanced,
    breakable,
    colback=black!2,
    colframe=black!35,
    boxrule=0.5pt,
    arc=1.5mm,
    left=2mm,
    right=2mm,
    top=1.5mm,
    bottom=1.5mm,
    fonttitle=\bfseries,
    coltitle=black,
    title=#1,
    before skip=6pt,
    after skip=6pt
}

\title{Rethinking Next-Visit Diagnosis Prediction Beyond Aggregate Metrics: \\ Retrieval–Memory Multi-Agent Adjudication}

\author{Heejoon Koo \\
  University of Illinois Urbana-Champaign \\
  \texttt{heejoon4@illinois.edu}}

\begin{document}
\maketitle

\begin{abstract}
Next-visit diagnosis prediction forecasts the diagnoses documented at a patient’s next encounter, turning retrospective electronic health records into prospective clinical prediction. However, conventional aggregate evaluation obscures its multifaceted nature: distributionally, models must handle variable-size diagnosis sets and long-tailed labels; temporally, they must distinguish persistent, newly recorded, and recurrent diagnoses; and both vary with observed history depth. We introduce an evaluation framework along these dimensions and apply it to rule-based heuristics, LLM-based prompting, retrieval, memory, and multi-agent approaches. This reveals that repeating the last visit outperforms most LLM-based approaches overall despite predicting no new or recurrent diagnoses, and that retrieval and memory exhibit distinct profiles whose relative strengths vary across temporal transitions and backbones. Motivated by this, we propose LONGINUS, a retrieval–memory multi-agent adjudication framework combining evidence retrieved from patient records with structured memory of the patient trajectory. Using cross-view agreement as a reliability heuristic, it preserves shared predictions and adjudicates disagreements without revealing their source. Across MIMIC-III and MIMIC-IV with four LLM backbones, LONGINUS achieves the best transition-specific performance among LLM-based methods in nearly all settings while remaining competitive with them on aggregate metrics.
\end{abstract}

\section{Introduction}

Electronic health records (EHRs) accumulate a patient's clinical history across encounters, thereby enabling a shift from retrospective documentation towards prospective clinical prediction. Next-visit diagnosis prediction formulates this objective as multi-label prediction over a sequence of prior visits: given a patient's record up to the current encounter, predict the set of diagnoses documented at the next visit \cite{choi2016doctor, li2020behrt, koo2024next}. Anticipating these diagnoses could facilitate proactive scheduling, preventive testing, and personalized intervention \cite{escobar2020automated, goldstein2020development, adams2022prospective}.

\begin{figure}[t]
    \centering
    \includegraphics[width=0.48\textwidth]{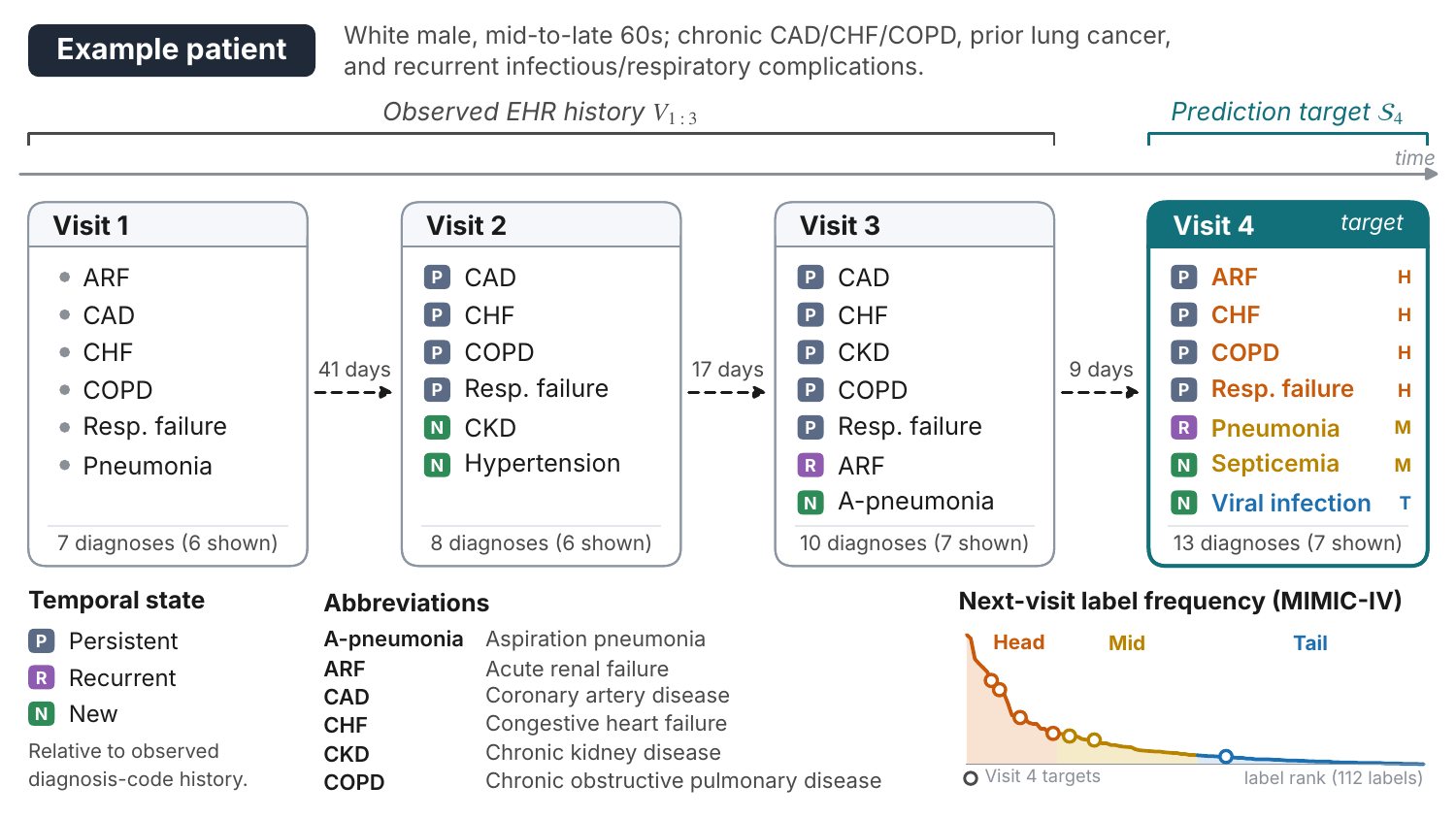}
    \caption{A synthetic illustrative example of the multifaceted nature of next-visit diagnosis prediction.}
    \label{fig:overview_longitudinal_patient_data}
    \vspace{-3mm}
\end{figure}

This formulation poses challenges that adjacent tasks do not. Single-label clinical prediction assigns one target per observation \cite{choi2016retain, baldenweg2024multi}; single-visit classification uses one encounter \cite{rajpurkar2017chexnet, baltruschat2019comparison}; and retrospective ICD coding labels an already documented encounter \cite{mullenbach2018explainable, huang2022plm}. In contrast, next-visit diagnosis prediction forecasts a diagnosis set from multi-visit history for a future encounter, combining distributional and temporal challenges over longitudinal histories (Figure~\ref{fig:overview_longitudinal_patient_data}). Distributionally, encounters contain varying numbers of co-occurring diagnoses drawn from a long-tailed label distribution. Temporally, diagnoses may persist, be newly coded relative to observed history, or reappear after an interval of absence. Both challenges also vary with history depth: label and transition distributions shift as visits accumulate, and recurrence requires at least two observed visits. Aggregate scores alone may therefore conceal performance differences across distributional settings, temporal transitions, and history depths.

\begin{table}[t]
\centering
\scriptsize
\setlength{\tabcolsep}{2.5pt}
\renewcommand{\arraystretch}{1.05}
\providecommand{\yes}{\ding{52}}
\providecommand{\no}{\textcolor{gray}{\ding{56}}}

\resizebox{\columnwidth}{!}{%
\begin{tabular}{@{}lccccccc@{}}
\toprule
\midrule
\multirow{2}{*}{\textbf{Prior Work}} &
\textbf{Overall} &
\multicolumn{2}{c}{\textbf{Distribution}} &
\multicolumn{3}{c}{\textbf{Temporal Transitions}} &
\multirow{2}{*}{\shortstack{\textbf{History}\\\textbf{Depth}}} \\
\cmidrule(lr){2-2}
\cmidrule(lr){3-4}
\cmidrule(lr){5-7}
& \textbf{Agg.} &
\textbf{Co} & \textbf{LT} &
\textbf{Pers.} & \textbf{New} & \textbf{Rec.} & \\
\hline
\midrule
GRAM \cite{choi2017gram}
& \yes & \no & \yes & \no & \no & \no & \no \\
TCoN \cite{sun2021interpretable}
& \yes & \yes & \no & \no & \no & \no & \no \\
NECHO \cite{koo2024next}
& \yes & \no & \no & \no & \no & \no & \no \\
TAKECare \cite{wu2025takecare}
& \yes & \yes & \no & \no & \no & \no & \yes \\
RAVEN \cite{rajamohan2026scaling}
& \yes & \no & \no & \no & \yes & \no & \no \\
\midrule
\textbf{Ours (Evaluation)}
& \yes & \yes & \yes & \yes & \yes & \yes & \yes \\
\bottomrule
\end{tabular}%
}

\caption{
Evaluation coverage in representative next-visit
diagnosis prediction studies.
Agg.: aggregate metrics;
Co: co-occurrence;
LT: long-tail;
Pers.: persistent;
New: newly recorded;
Rec.: recurrent.
\yes/\no: explicitly evaluated/not explicitly evaluated.
}
\label{tab:literature_dimensions}
\vspace{-3mm}
\end{table}

Earlier approaches to next-visit diagnosis prediction modeled visit sequences using recurrent and attention-based networks \cite{choi2016doctor, choi2016retain, ma2017dipole}, medical ontologies \cite{choi2017gram}, and transformer-based or multimodal architectures \cite{peng2021sequential, li2020behrt, koo2024next}. More recently, LLM-based methods have explored ontology-aware hierarchical Chain-of-Thought (CoT) prompting \cite{koo2025towards} and fine-tuning \cite{ben2024cpllm}.

The evaluation of these diverse approaches, by contrast, remains fragmented (Table~\ref{tab:literature_dimensions}). Distributional analyses have examined within-visit co-occurrence \cite{sun2021interpretable} and low-frequency diagnoses \cite{choi2017gram}. Temporal analyses have focused primarily on newly appearing diagnoses: RAVEN emphasizes new onsets by downweighting repeated events \cite{rajamohan2026scaling}. EviCare also targets novel diagnoses \cite{zhang2026evicare}. History depth has been examined separately \cite{wu2025takecare}, and \citet{kruse2025large} show that LLMs struggle with both rare diagnoses and temporal progression. Nevertheless, to our knowledge, no prior study jointly evaluates distributional factors (multi-label set performance and label frequency), temporal transitions (persistent, new, and recurrent diagnoses), and history depth, leaving unclear how model strengths and limitations vary across these dimensions (see Appendix \ref{app:related_work}).

We therefore introduce a unified, model-agnostic, distribution- and transition-aware evaluation framework that jointly assesses multi-label set performance, label frequency, temporal transitions, and observed history depth. On MIMIC-IV \cite{johnson2023mimic}, this framework reveals that aggregate scores can reward diagnostic persistence: last-visit carry-forward (LVCF) outperforms most LLM-based results in Micro-F1 (21 of 24; Section~\ref{sec:preliminary}) despite predicting no New or Recurrent diagnoses, while observed-history union (OHU) can predict Recurrent but not New diagnoses. It also distinguishes the retrieval and memory views: retrieval achieves higher Micro-F1 across all backbones and history depths, whereas memory achieves higher Recurrent F1 with three of four backbones and higher New F1 in most settings, while Persistent F1 shows no consistent direction. These distinct profiles motivate reconciling the two views with patient-specific evidence, rather than relying on either view alone.

We propose \textbf{LONGINUS} (\textbf{LONG}itudinal \textbf{IN}tegration of retrieval and memory \textbf{U}sing
\textbf{S}ource-neutral adjudication), a retrieval--memory disagreement-guided multi-agent adjudication framework that generates separate predictions from evidence retrieved from the patient's record and a structured memory of the patient trajectory. Using cross-view agreement as a reliability heuristic, LONGINUS preserves shared predictions and reserves adjudication for candidates proposed by only one view. The adjudicator reviews these candidates using both evidence views and diagnosis-history states, without revealing which view proposed each. This design encourages evidence-based decisions without imposing an explicit preference for either retrieval or memory. In summary, our contributions are threefold:
\begin{itemize}[leftmargin=*, noitemsep, topsep=0pt]
\item We introduce a model-agnostic, distribution- and transition-aware evaluation framework that assesses multi-label set performance and label frequency, decomposes predictions into persistent, new, and recurrent diagnoses, and stratifies results by observed history depth.
\item We benchmark training-free LLM-based prompting, retrieval, memory, and multi-agent approaches, showing that LVCF exceeds every LLM-based method in Micro-F1 on MIMIC-III despite predicting no new or recurrent diagnoses, and that retrieval and memory exhibit distinct prediction profiles across transitions, backbones, and history depths.
\item We propose \textbf{LONGINUS}, a training-free, LLM-based multi-agent framework that combines local retrieval and global trajectory memory via consensus preservation and source-neutral adjudication. Across two datasets and four backbones, it achieves the best transition-specific F1 among training-free LLM-based methods in 22 of 24 settings while remaining competitive with them on aggregate metrics. Ablation, comparative, and stage-wise analyses further support our design choices.
\end{itemize}

\section{Preliminary Analysis}
\label{sec:preliminary}

\subsection{Problem Formulation}

For a patient $p$, let $V_{1:T}=(V_1,\ldots,V_T)$ denote the observed longitudinal record, where each visit $V_t=(D_t, N_t, \mathcal{S}_t)$ consists of demographic information $D_t$, a clinical note $N_t$, and the set of documented diagnoses $\mathcal{S}_t \subseteq \mathcal{L}$ over the fixed label space $\mathcal{L}$. The objective is to predict the diagnosis set at the subsequent visit,
{
\setlength{\abovedisplayskip}{8pt}
\setlength{\belowdisplayskip}{8pt}
\begin{equation}
\hat{\mathcal{S}}_{T+1} = f(V_{1:T}) \subseteq \mathcal{L},
\end{equation}
}
where $f$ denotes a prediction model, including rule-based heuristics and LLM-based approaches. We stratify patients by observed history depth: one observed visit (Visit 1) or two or more (Visit 2+). This split separates histories in which recurrence is identifiable from those in which it is not. 

\subsection{Temporal Decomposition by Visit History}

For an observed visit $t \le T$, let $\mathcal{H}_t=\bigcup_{\tau \le t}\mathcal{S}_{\tau}$ denote the diagnostic history up to and including visit $t$. We decompose the next-visit diagnoses $\mathcal{S}_{t+1}$ by their relationship to this history:
\begin{align}
\mathcal{S}^{\mathrm{pers}}_{t+1} &= \mathcal{S}_{t+1} \cap \mathcal{S}_t,\\
\mathcal{S}^{\mathrm{new}}_{t+1}  &= \mathcal{S}_{t+1} \setminus \mathcal{H}_{t},\\
\mathcal{S}^{\mathrm{rec}}_{t+1}  &= \mathcal{S}_{t+1} \cap (\mathcal{H}_{t}\setminus\mathcal{S}_t),
\end{align}
where $\mathcal{S}^{\mathrm{pers}}_{t+1}$, $\mathcal{S}^{\mathrm{new}}_{t+1}$, and $\mathcal{S}^{\mathrm{rec}}_{t+1}$ denote Persistent, New, and Recurrent diagnoses, respectively. These sets partition $\mathcal{S}_{t+1}$; New is defined relative to the observed coding history and may include newly developed conditions, previously uncoded comorbidities, or newly documented complications. With one observed visit, $\mathcal{S}^{\text{rec}}_{t+1}$ is empty by construction, so Recurrent F1 is reported only for Visit~2+. In Visit~2+, Recurrent diagnoses emerge and the share of New diagnoses drops, while Persistent diagnoses remain the largest component (Table~\ref{tab:preliminary_transition_distribution}).

\subsection{What Aggregate Metrics Hide}

In Table~\ref{tab:preliminary_performance}, LVCF exceeds 21 of 24 LLM-based results in Micro-F1 despite predicting no New or Recurrent diagnoses and OHU also predicts no New diagnoses\footnote{Patterns were first observed on the development cohort and are reported here on the evaluation cohort.}. Even when NECHOv3 with Llama-3.3-70B exceeds LVCF in Visit~2+, its New F1 is only .082. Since Persistent diagnoses dominate next-visit labels (Table~\ref{tab:preliminary_transition_distribution}), aggregate scores can favor persistence over predicting previously unobserved diagnoses.

\begin{table}[h]
\centering
\scriptsize
\setlength{\tabcolsep}{5pt}
\renewcommand{\arraystretch}{1.05}
\begin{tabular}{@{}cccccc@{}}
\toprule
\midrule
\textbf{Dataset} &
\textbf{History} &
\textbf{Patients (\%)} &
\textbf{New} &
\textbf{Persistent} &
\textbf{Recurrent} \\
\hline
\midrule
\multirow{2}{*}{MIMIC-III}
& Visit 1  & 78.6 & 45.64 & 54.36 & -- \\
& Visit 2+ & 21.4 & 27.52 & 59.31 & 13.17 \\
\midrule
\multirow{2}{*}{MIMIC-IV}
& Visit 1  & 74.4 & 41.37 & 58.63 & -- \\
& Visit 2+ & 25.6 & 25.05 & 59.14 & 15.81 \\
\bottomrule
\end{tabular}
\caption{Temporal composition (\%) of next-visit diagnosis labels by observed history depth. Patients: share of patients per dataset; diagnosis percentages are over all target labels in each history group. Recurrent is undefined for Visit 1 (--).}
\vspace{-3mm}
\label{tab:preliminary_transition_distribution}
\end{table}

\begin{table}[h]
\centering
\scriptsize
\setlength{\tabcolsep}{3.6pt}
\renewcommand{\arraystretch}{1.0}

\begin{tabular}{ccccc|cccc}
\toprule
\midrule
\multirow{3}{*}{\textbf{Backbone}}
& \multirow{3}{*}{\textbf{Method}}
& \multicolumn{3}{c|}{\textbf{Visit 1}}
& \multicolumn{4}{c}{\textbf{Visit 2+}} \\
\cmidrule(lr){3-5}\cmidrule(lr){6-9}

&& \textbf{Agg.}
& \multicolumn{2}{c|}{\textbf{Temp.}}
& \textbf{Agg.}
& \multicolumn{3}{c}{\textbf{Temp.}} \\
\cmidrule(lr){3-3}\cmidrule(lr){4-5}
\cmidrule(lr){6-6}\cmidrule(lr){7-9}

&& \textbf{Mic.}
& \textbf{Pers.}
& \textbf{New}
& \textbf{Mic.}
& \textbf{Pers.}
& \textbf{New}
& \textbf{Rec.} \\

\hline
\midrule

\multirow{2}{*}{Rule}
& LVCF
& .582 & .732 & .000
& .586 & .734 & .000 & .000 \\

& OHU
& .582 & .732 & .000
& .582 & .734 & .000 & .444 \\
\midrule

\multirow{3}{*}{\makecell{Gemini-2.5\\-Flash-Lite}}
& NECHOv3
& \belowrule{.493}
& \belowrule{.722}
& .203
& \belowrule{\underline{.521}}
& \underline{.741}
& .148
& \underline{.484} \\

& EHR-RAG
& \belowrule{\textbf{.564}}
& \belowrule{\underline{.731}}
& \underline{.247}
& \belowrule{\textbf{.574}}
& .737
& \underline{.157}
& \belowrule{.437} \\

& Traj-CoA
& \belowrule{\underline{.531}}
& \textbf{.741}
& \textbf{.283}
& \belowrule{.507}
& \textbf{.743}
& \textbf{.186}
& \textbf{.495} \\
\midrule

\multirow{3}{*}{\makecell{Gemini-3.1\\-Flash-Lite}}
& NECHOv3
& \belowrule{\underline{.552}}
& \textbf{.752}
& \underline{.254}
& \belowrule{\underline{.552}}
& \textbf{.768}
& .153
& .453 \\

& EHR-RAG
& \textbf{.584}
& \underline{.748}
& \textbf{.315}
& \belowrule{\textbf{.575}}
& \underline{.753}
& \textbf{.213}
& \textbf{.488} \\

& Traj-CoA
& \belowrule{.526}
& \belowrule{.719}
& .250
& \belowrule{.525}
& .735
& \underline{.182}
& \underline{.455} \\
\midrule

\multirow{3}{*}{\makecell{GPT-OSS\\-120B}}
& NECHOv3
& \belowrule{.564}
& \belowrule{.722}
& \underline{.290}
& \belowrule{\underline{.561}}
& \belowrule{.713}
& \textbf{.223}
& \underline{.472} \\

& EHR-RAG
& \belowrule{\textbf{.579}}
& \underline{.732}
& .265
& \textbf{.589}
& \textbf{.743}
& .177
& .450 \\

& Traj-CoA
& \belowrule{\underline{.571}}
& \textbf{.748}
& \textbf{.303}
& \belowrule{.545}
& \underline{.740}
& \underline{.206}
& \textbf{.483} \\
\midrule

\multirow{3}{*}{\makecell{Llama-3.3\\-70B}}
& NECHOv3
& \belowrule{\underline{.533}}
& \belowrule{\underline{.710}}
& .064
& \textbf{.600}
& \textbf{.747}
& .082
& \belowrule{\underline{.406}} \\

& EHR-RAG
& \belowrule{\textbf{.548}}
& \belowrule{\textbf{.713}}
& \underline{.138}
& \belowrule{\underline{.584}}
& \belowrule{\underline{.731}}
& \underline{.107}
& \belowrule{.350} \\

& Traj-CoA
& \belowrule{.494}
& \belowrule{.656}
& \textbf{.227}
& \belowrule{.527}
& \belowrule{.686}
& \textbf{.184}
& \textbf{.453} \\

\bottomrule
\end{tabular}

\caption{Performance on MIMIC-IV by observed history depth. NECHOv3 is a single-prompt, full-context baseline \cite{koo2025towards}; EHR-RAG \cite{cao2026ehr} and Traj-CoA \cite{zeng2025trajcoa} are representative patient-internal retrieval and trajectory-memory baselines, respectively (Section~\ref{sec:baselines}). Agg.: Micro-F1; Temp.: transition-specific F1 (Rec. in Visit 2+ only). \textbf{Bold}/\underline{underline}: best/second-best LLM-based result per backbone and column using unrounded values; Gray: below the best rule-based score per column.}
\vspace{-3mm}
\label{tab:preliminary_performance}
\end{table}

\subsection{Retrieval--Memory Prediction Profiles}

Existing retrieval- and memory-based methods show distinct profiles (Table~\ref{tab:preliminary_performance}): across all four backbones, EHR-RAG achieves higher Micro-F1 than Traj-CoA in both history strata, whereas Traj-CoA achieves higher Recurrent F1 with three of four backbones and higher New F1 in six of eight backbone--depth settings. These strengths are backbone-dependent: Persistent F1 shows no consistent direction, and with Gemini-3.1-Flash-Lite, EHR-RAG leads on every transition. Since neither view dominates, these profiles motivate combining the two without a fixed preference: LONGINUS builds history-state-aware retrieval and structured memory views, retains shared predictions, and applies source-neutral review to disagreements (Section~\ref{sec:methodology}). 

\begin{figure*}[h]
    \centering
    \includegraphics[width=0.9\textwidth]{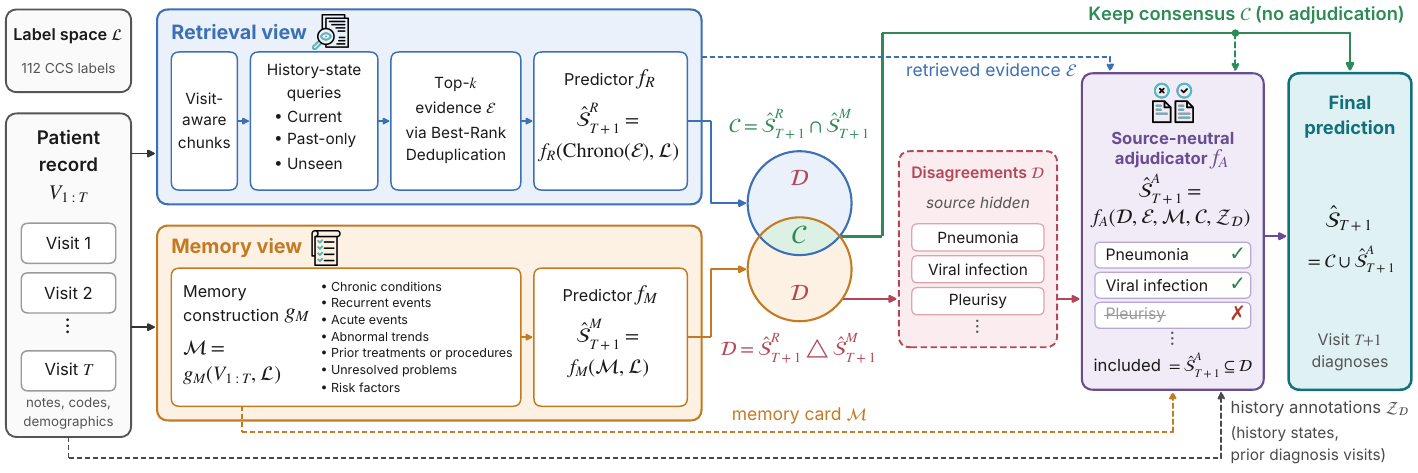}
    \caption{Overview of Our LONGINUS Framework. Candidate labels follow the example patient in Figure \ref{fig:overview_longitudinal_patient_data}.}
    \label{fig:overview_longinus_framework}
    \vspace{-3mm}
\end{figure*}

\section{Methodology}
\label{sec:methodology}

LONGINUS instantiates $f$ with three components: a retrieval view that selects local evidence from the patient's record, a memory view that summarizes the global patient trajectory, and a source-neutral adjudicator that reviews disagreements between their predictions while retaining shared labels.

\subsection{Retrieval View}
\label{sec:methodology_retrieval_view}

The retrieval view selects local evidence from the patient's observed EHR trajectory. We partition the fixed label space $\mathcal{L}$ according to diagnostic history:
{
\setlength{\abovedisplayskip}{8pt}
\setlength{\belowdisplayskip}{8pt}
\begin{align}
\mathcal{L}_{\mathrm{cur}} &= \mathcal{S}_T,
&& \text{Current},\\
\mathcal{L}_{\mathrm{past}} &= \mathcal{H}_{T-1}\setminus\mathcal{S}_T,
&& \text{Past-only},\\
\mathcal{L}_{\mathrm{uns}} &= \mathcal{L}\setminus\mathcal{H}_T,
&& \text{Unseen}.
\end{align}
}
These history states are observable at prediction time, whereas the Persistent, Recurrent, and New transitions in Section~\ref{sec:preliminary} depend on the next-visit outcome. They guide retrieval and later annotate adjudication candidates.

We perform patient-internal dense retrieval over visit-aware EHR chunks using one query per history state. The Current query seeks evidence of continued activity or resolution; the Past-only query targets prior episodes and possible reactivation; and the Unseen query seeks symptoms, abnormal findings, diagnostic workup, treatments, or complications that may support diagnoses not previously documented in the observed history.

Let $r_q(c)$ denote the rank of chunk $c$ under retrieval path $q\in\{\mathrm{cur},\mathrm{past},\mathrm{uns}\}$, with $r_q(c)=\infty$ if that path does not retrieve $c$. We merge the retrieved chunks, remove duplicates, and assign each unique chunk its best rank:
{
\setlength{\abovedisplayskip}{8pt}
\setlength{\belowdisplayskip}{8pt}
\begin{equation}
r^{*}(c)=\min_q r_q(c).
\end{equation}
}
We select the $k$ unique chunks with the best ranks (ties broken by similarity score), then present these selected chunks to the LLM in chronological order to preserve the temporal sequence of the retrieved evidence,
\begin{equation}
\begin{aligned}
\mathcal{E}
&= \operatorname{TopK}_{c \in \mathcal{U}}
   \left(-r^{*}(c)\right),\\
\hat{\mathcal{S}}_{T+1}^{R}
&= f_R\left(\operatorname{Chrono}(\mathcal{E}),
   \mathcal{L}\right),
\end{aligned}
\end{equation}
where $\mathcal{U}$ is the set of unique chunks retrieved across all paths, $\operatorname{TopK}$ selects the $k=12$ highest-scoring chunks, and $\operatorname{Chrono}$ orders them by visit time. Best-rank deduplication (BRD) prioritizes evidence ranked highly by any path without rewarding repeated retrieval across paths.

\subsection{Memory View}

The memory view summarizes the observed patient trajectory through two LLM calls. The first summarizes the complete record into a structured memory card covering chronic conditions, recurrent and acute events, abnormal trends, prior treatments or procedures, unresolved problems, and risk factors. The second predicts diagnoses using this card and the fixed label space:
{
\setlength{\abovedisplayskip}{8pt}
\setlength{\belowdisplayskip}{8pt}
\begin{equation}
\begin{aligned}
\mathcal{M}
&= g_M(V_{1:T}, \mathcal{L}),\\
\hat{\mathcal{S}}_{T+1}^{M}
&= f_M(\mathcal{M}, \mathcal{L}),
\end{aligned}
\end{equation}
}
where $g_M$ constructs the structured clinical memory $\mathcal{M}$
and $f_M$ produces the memory-view prediction.

\subsection{Source-Neutral Disagreement Adjudication}

Given the separately generated predictions $\hat{\mathcal{S}}_{T+1}^{R}$ and $\hat{\mathcal{S}}_{T+1}^{M}$, we partition their union into consensus and disagreement:
{
\setlength{\abovedisplayskip}{8pt}
\setlength{\belowdisplayskip}{8pt}
\begin{equation}
\begin{aligned}
\mathcal{C}
&= \hat{\mathcal{S}}_{T+1}^{R}
   \cap \hat{\mathcal{S}}_{T+1}^{M},\\
\mathcal{D}
&= \hat{\mathcal{S}}_{T+1}^{R}
   \triangle \hat{\mathcal{S}}_{T+1}^{M},
\end{aligned}
\end{equation}
}
where $\triangle$ denotes symmetric difference. Treating cross-view agreement as a signal of reliability, we retain labels in $\mathcal{C}$ and forward only $\mathcal{D}$ to the adjudicator $f_A$. Each candidate includes its history state (Current, Past-only, or Unseen) and prior diagnosis visits, if any; these annotations are denoted collectively by $\mathcal{Z}_{\mathcal{D}}$.

To discourage source-based preferences, $f_A$ is not told which view proposed each label. A single batched LLM call decides each candidate with history-state-specific criteria: continued activity for Current, no exclusion solely for absence from the latest visit for Past-only, and patient-specific support for Unseen. Given $\mathcal{E}$, $\mathcal{M}$, $\mathcal{C}$, and $\mathcal{Z}_\mathcal{D}$, the adjudicator returns an accepted subset, combining with the consensus to form the final prediction:
{
\setlength{\abovedisplayskip}{8pt}
\setlength{\belowdisplayskip}{8pt}
\begin{equation}
\begin{aligned}
\hat{\mathcal{S}}_{T+1}^{A}
&= f_A\left(
\mathcal{D}, \mathcal{E}, \mathcal{M},
\mathcal{C}, \mathcal{Z}_{\mathcal{D}}
\right)
\subseteq \mathcal{D},\\
\hat{\mathcal{S}}_{T+1}
&= \mathcal{C} \cup \hat{\mathcal{S}}_{T+1}^{A}.
\end{aligned}
\end{equation}
}

\section{Experiments}
\subsection{Experimental Setup}

\subsubsection{Data and Preprocessing}
We construct MIMIC-III \cite{johnson2016mimic} and MIMIC-IV
\cite{johnson2023mimic} cohorts from discharge summaries, diagnosis
codes, and demographics following prior protocols \cite{choi2016doctor, koo2024next}. We retain adults with valid urgent, emergency, or elective admissions and, after predefined exclusions, map diagnoses to 112 CCS single-level labels, which form the label space $\mathcal{L}$ and are grouped into 14 multi-level parent categories. Each prediction uses up to five chronologically ordered visits immediately preceding the target, with consecutive gaps of at most 180 days, including the observed-to-target gap \cite{weiskopf2013defining}. We sample 1,000 unique patients per dataset, stratified by age group, sex, race, and visit depth. Details appear in Appendix~\ref{app:preprocessing}.

\subsubsection{Evaluation Metrics}
We report Micro-F1 for aggregate performance, patient-averaged Jaccard for set overlap, and label-averaged Macro-F1 for infrequent diagnoses \cite{wang2026integrating}, after removing out-of-space predictions. Transition-specific F1 maps the history states of Section~\ref{sec:methodology} (Current, Unseen, Past-only) to Persistent, New, and Recurrent and applies them to both predicted and reference labels, so that false positives are also attributed to a transition. All metrics are further stratified by observed history depth; formal definitions are given in Appendix~\ref{app:eval_metrics}.

\subsubsection{LLM Backbones}
We evaluate four LLM backbones: Gemini-2.5-Flash-Lite \cite{comanici2025gemini}, Gemini-3.1-Flash-Lite \cite{team2023gemini, googledeepmind2026gemini31flashlite}, GPT-OSS-120B \cite{agarwal2025gpt}, and Llama-3.3-70B \cite{grattafiori2024llama}. Section~\ref{sec:preliminary} reports all four; Section~4 presents detailed results for Gemini-3.1-Flash-Lite and Llama-3.3-70B, with the remaining two in Appendix~\ref{app:appendix_full_results}. Given inconsistent gains from medical adaptation \cite{jeong2024medical} and backbone sensitivity \cite{zhuo2024prosa,zhou2024larger}, we evaluate
general-purpose LLMs across model families, holding the backbone fixed within each setting.

\subsubsection{Baselines}
\label{sec:baselines}
We compare LONGINUS with rule-based heuristics (LVCF and OHU) and four families of LLM-based methods: single-prompting (NECHOv3 \cite{koo2025towards}), multi-agent (MedAgents \cite{tang2024medagents}, MDAgents \cite{kim2024mdagents}, and KAMAC \cite{wu2025knowledge}), retrieval (Summary-RAG \cite{xiong2024benchmarking}, RGAR \cite{liang2025rgar}, MASS-RAG \cite{xiao2026mass}, and EHR-RAG \cite{cao2026ehr}), and memory (Traj-CoA \cite{zeng2025trajcoa}), with task-specific adaptations described in Appendix \ref{app:baselines}.

We focus on training-free LLM methods that require no task-specific training or retraining for new cohorts. Where feasible, we standardize system and task prompts, label space, output format, and recall-oriented guidance \cite{koo2025towards}, while preserving method-specific procedures.

\subsubsection{Implementation Details}
\label{sec:implementation}
We use temperature 0 for all LLM calls. Dense retrieval uses \texttt{bge-small-en-v1.5} \cite{xiao2024c}. LONGINUS and patient-internal RAG baselines use 160-token chunks with 40-token overlap and retrieve the top 12 chunks. Full prompts of LONGINUS are provided in Appendix~\ref{app:prompts}. Design choices, including the adjudication variant (Section~\ref{sec:comparative_studies}), were selected on a separate patient-disjoint development cohort of 500 patients from MIMIC-III and applied unchanged to MIMIC-IV, prioritizing transition-specific F1; the 1,000-patient cohorts were reserved exclusively for final evaluation. We assess statistical significance with two-sided paired patient-level bootstrap tests (2000 replicates; $p<.05$), applying Holm correction within each comparison family.

\begin{table*}[t]
\centering
\scriptsize
\setlength{\tabcolsep}{5.0pt}
\renewcommand{\arraystretch}{1.1}

\begin{tabular}{@{}ccccccccccccccc@{}}
\toprule
\midrule
\multirow{3}{*}{\textbf{Backbone}} &
\multirow{3}{*}{\textbf{Type}} &
\multirow{3}{*}{\textbf{Method}} &
\multicolumn{6}{c}{\textbf{MIMIC-III}} &
\multicolumn{6}{c}{\textbf{MIMIC-IV}} \\
\cmidrule(lr){4-9}\cmidrule(lr){10-15}
& & &
\textbf{Agg.} & \multicolumn{2}{c}{\textbf{Dist.}} &
\multicolumn{3}{c}{\textbf{Temp.}} &
\textbf{Agg.} & \multicolumn{2}{c}{\textbf{Dist.}} &
\multicolumn{3}{c}{\textbf{Temp.}} \\
\cmidrule(lr){4-4}\cmidrule(lr){5-6}\cmidrule(lr){7-9}
\cmidrule(lr){10-10}\cmidrule(lr){11-12}\cmidrule(lr){13-15}
& & &
\textbf{Mic.} & \textbf{Mac.} & \textbf{Jac.} &
\textbf{Pers.} & \textbf{New} & \textbf{Rec.} &
\textbf{Mic.} & \textbf{Mac.} & \textbf{Jac.} &
\textbf{Pers.} & \textbf{New} & \textbf{Rec.} \\
\hline
\midrule

\multirow{2}{*}{-} & \multirow{2}{*}{Rule} & LVCF
& .558 & .510 & .397 & .719 & .000 & .000
& .582 & .479 & .439 & .732 & .000 & .000 \\

& & OHU
& .553 & .510 & .391 & .719 & .000 & .383
& .582 & .492 & .436 & .732 & .000 & .444 \\

\midrule

\multirow{10}{*}{\makecell{Gemini-3.1\\-Flash-Lite}}
& Single-Prompting & NECHOv3
& \belowrule{.514} & \belowrule{.450} & \belowrule{.353}
& \belowrule{.715} & .194 & .393
& \belowrule{.552} & \belowrule{.459} & \belowrule{.392}
& \underline{.756} & .236 & .453 \\

\cmidrule(lr){2-15}

& \multirow{3}{*}{Multi-agent} & MedAgents
& \belowrule{.506} & \belowrule{.407} & \belowrule{.347}
& \belowrule{.695} & .217 & \belowrule{.383}
& \belowrule{.547} & \belowrule{.444} & \belowrule{.390}
& .736 & .264 & \belowrule{.416} \\

& & MDAgents
& \belowrule{.505} & \belowrule{.443} & \belowrule{.341}
& \belowrule{.700} & .210 & \belowrule{.368}
& \belowrule{.546} & \belowrule{.474} & \belowrule{.387}
& .736 & .263 & .478 \\

& & KAMAC
& \belowrule{.485} & \belowrule{.379} & \belowrule{.324}
& \belowrule{.670} & .200 & \belowrule{.372}
& \belowrule{.525} & \belowrule{.422} & \belowrule{.365}
& \belowrule{.703} & .270 & \belowrule{.435} \\

\cmidrule(lr){2-15}

& \multirow{4}{*}{Retrieval} & Summary-RAG
& \belowrule{.487} & \belowrule{.405} & \belowrule{.330}
& \belowrule{.693} & .222 & \underline{.409}
& \belowrule{.518} & \belowrule{.437} & \belowrule{.356}
& \belowrule{.728} & .267 & \underline{.491} \\

& & RGAR
& \belowrule{.472} & \belowrule{.391} & \belowrule{.318}
& \belowrule{.674} & .196 & \belowrule{.374}
& \belowrule{.512} & \belowrule{.426} & \belowrule{.359}
& \belowrule{.705} & .254 & .473 \\

& & MASS-RAG
& \belowrule{.526} & \belowrule{.454} & \belowrule{\underline{.368}}
& \belowrule{.713} & .193 & .383
& \belowrule{.578} & \belowrule{.482} & \belowrule{.418}
& .746 & .265 & .487 \\

& & EHR-RAG
& \belowrule{\underline{.530}} & \belowrule{\underline{.460}}
& \belowrule{.367}
& \belowrule{\underline{.716}} & \underline{.224} & .383
& \belowrule{\underline{.582}} & \belowrule{\underline{.484}}
& \belowrule{\textbf{.420}}
& .749 & \underline{.298} & .488 \\

\cmidrule(lr){2-15}

& Memory & Traj-CoA
& \belowrule{.484} & \belowrule{.402} & \belowrule{.332}
& \belowrule{.691} & .184 & \belowrule{.372}
& \belowrule{.525} & \belowrule{.428} & \belowrule{.380}
& \belowrule{.722} & .237 & .455 \\

\cmidrule(lr){2-15}

& Ours & \textbf{LONGINUS}
& \belowrule{\textbf{.535}} & \belowrule{\textbf{.462}}
& \belowrule{\textbf{.370}}
& \textbf{.728}$^{\dagger}$ & \textbf{.255}$^{\dagger}$ & \textbf{.439}
& \textbf{.583} & \belowrule{\textbf{.490}} & \belowrule{\underline{.418}}
& \textbf{.773}$^{\dagger}$ & \textbf{.314} & \textbf{.529} \\

\midrule

\multirow{10}{*}{\makecell{Llama-3.3\\-70B}}
& Single-Prompting & NECHOv3
& \belowrule{.510} & \belowrule{.405} & \belowrule{.354}
& \belowrule{.686} & .021 & \belowrule{.297}
& \belowrule{.548} & \belowrule{.405} & \belowrule{.401}
& \belowrule{.719} & .067 & \belowrule{.406} \\

\cmidrule(lr){2-15}

& \multirow{3}{*}{Multi-agent} & MedAgents
& \belowrule{.428} & \belowrule{.302} & \belowrule{.284}
& \belowrule{.606} & .109 & \belowrule{.241}
& \belowrule{.468} & \belowrule{.325} & \belowrule{.330}
& \belowrule{.630} & .167 & \belowrule{.378} \\

& & MDAgents
& \belowrule{.449} & \belowrule{.328} & \belowrule{.303}
& \belowrule{.632} & .105 & \belowrule{.248}
& \belowrule{.487} & \belowrule{.352} & \belowrule{.348}
& \belowrule{.656} & .170 & \belowrule{.344} \\

& & KAMAC
& \belowrule{.465} & \belowrule{.357} & \belowrule{.315}
& \belowrule{.643} & .112 & \belowrule{.250}
& \belowrule{.502} & \belowrule{.389} & \belowrule{.362}
& \belowrule{.668} & .184 & \belowrule{.358} \\

\cmidrule(lr){2-15}

& \multirow{4}{*}{Retrieval} & Summary-RAG
& \belowrule{.390} & \belowrule{.293} & \belowrule{.261}
& \belowrule{.557} & .127 & \belowrule{.268}
& \belowrule{.484} & \belowrule{.378} & \belowrule{.342}
& \belowrule{.676} & \underline{.226} & \belowrule{.436} \\

& & RGAR
& \belowrule{.463} & \belowrule{.358} & \belowrule{.321}
& \belowrule{.642} & .056 & \belowrule{.196}
& \belowrule{.491} & \belowrule{.372} & \belowrule{.370}
& \belowrule{.657} & .127 & \belowrule{.269} \\

& & MASS-RAG
& \belowrule{\underline{.518}} & \belowrule{.434}
& \belowrule{\underline{.360}}
& \belowrule{\underline{.694}} & .102
& \belowrule{\underline{.360}}
& \belowrule{\underline{.562}} & \belowrule{\underline{.455}}
& \belowrule{.415}
& \belowrule{\underline{.723}} & .195 & .450 \\

& & EHR-RAG
& \belowrule{.508} & \belowrule{\underline{.437}}
& \belowrule{.357}
& \belowrule{.683} & .082 & \belowrule{.160}
& \belowrule{.556} & \belowrule{.446}
& \belowrule{\underline{.425}}
& \belowrule{.717} & .134 & \belowrule{.350} \\

\cmidrule(lr){2-15}

& Memory & Traj-CoA
& \belowrule{.457} & \belowrule{.362} & \belowrule{.313}
& \belowrule{.632} & \underline{.139} & \belowrule{.345}
& \belowrule{.502} & \belowrule{.396} & \belowrule{.364}
& \belowrule{.663} & .220 & \underline{.453} \\

\cmidrule(lr){2-15}

& Ours & \textbf{LONGINUS}
& \belowrule{\textbf{.547}$^{\dagger}$}
& \belowrule{\textbf{.470}$^{\dagger}$}
& \belowrule{\textbf{.388}$^{\dagger}$}
& \textbf{.725}$^{\dagger}$
& \textbf{.174}$^{\dagger}$
& \belowrule{\textbf{.366}}
& \textbf{.597}$^{\dagger}$
& \belowrule{\textbf{.488}$^{\dagger}$}
& \textbf{.447}$^{\dagger}$
& \textbf{.759}$^{\dagger}$
& \textbf{.253}$^{\dagger}$
& \textbf{.481} \\

\bottomrule
\end{tabular}

\caption{Next-visit diagnosis prediction performance for two representative LLM backbones. \textbf{Bold}/\underline{underline}: best/second-best LLM-based result per backbone and column using unrounded values. Gray: below the best rule-based score in that column (rule-based baselines are backbone-independent). $\dagger$: LONGINUS significantly exceeds the highest-scoring competing LLM-based method in that column (Section~\ref{sec:implementation}).}
\vspace{-3mm}
\label{tab:main_results}
\end{table*}

\subsection{Overall and Temporal Performance}
\label{sec:overall_results}

\subsubsection{Main Results}
\label{sec:main_results}

Table~\ref{tab:main_results} reports results for Gemini-3.1-Flash-Lite and Llama-3.3-70B. Among LLM-based methods, LONGINUS achieves the best score in 23 of 24 dataset--backbone--metric combinations, including all 12 transition-specific ones; the only exception is MIMIC-IV Jaccard with Gemini-3.1-Flash-Lite, where EHR-RAG leads. Improvements are significant in 10 of 12 columns with Llama-3.3-70B, where LONGINUS exceeds the strongest LLM-based baseline on MIMIC-IV by .035, .027, and .028 in Micro-F1, New F1, and Recurrent F1, respectively, whereas with Gemini-3.1-Flash-Lite margins are smaller in aggregate metrics (e.g., .583 vs.\ .582 Micro-F1 on MIMIC-IV) than in Recurrent F1 (.529 vs.\ .491). Across all four backbones (Appendix~\ref{app:appendix_full_results}), LONGINUS attains the best transition-specific F1 in 22 of 24 settings; gains are consistent in direction but largest and most often significant with Llama-3.3-70B, and are achieved with at most four LLM calls per patient (Appendix~\ref{app:baselines}).

Rule-based methods nevertheless remain competitive in aggregate scores: on MIMIC-III, LVCF exceeds LONGINUS with both backbones despite predicting no New or Recurrent diagnoses. Since LONGINUS still exceeds the rule-based Persistent F1 in all four settings, this gap likely reflects imprecise non-persistent predictions. Appendix~\ref{app:appendix_full_results} reports full results across four backbones and label-frequency strata.

\begin{table*}[t]
\centering
\scriptsize
\setlength{\tabcolsep}{5pt}
\renewcommand{\arraystretch}{1.1}

\begin{tabular}{@{}cccccccccccccc@{}}
\toprule
\midrule
\multirow{3}{*}{\textbf{Backbone}} &
\multirow{3}{*}{\textbf{Type}} &
\multirow{3}{*}{\textbf{Method}} &
\multicolumn{5}{c}{\textbf{Visit 1}} &
\multicolumn{6}{c}{\textbf{Visit 2+}} \\
\cmidrule(lr){4-8}\cmidrule(lr){9-14}
& & &
\textbf{Agg.} & \multicolumn{2}{c}{\textbf{Dist.}} &
\multicolumn{2}{c}{\textbf{Temp.}} &
\textbf{Agg.} & \multicolumn{2}{c}{\textbf{Dist.}} &
\multicolumn{3}{c}{\textbf{Temp.}} \\
\cmidrule(lr){4-4}\cmidrule(lr){5-6}\cmidrule(lr){7-8}
\cmidrule(lr){9-9}\cmidrule(lr){10-11}\cmidrule(lr){12-14}
& & &
\textbf{Mic.} & \textbf{Mac.} & \textbf{Jac.} &
\textbf{Pers.} & \textbf{New} &
\textbf{Mic.} & \textbf{Mac.} & \textbf{Jac.} &
\textbf{Pers.} & \textbf{New} & \textbf{Rec.} \\
\hline
\midrule

\multirow{2}{*}{-} & \multirow{2}{*}{Rule} & LVCF
& .582 & .470 & .440 & .732 & .000
& .586 & .402 & .437 & .734 & .000 & .000 \\

& & OHU
& .582 & .470 & .440 & .732 & .000
& .582 & .455 & .426 & .734 & .000 & .444 \\

\midrule

\multirow{10}{*}{\makecell{Gemini-3.1\\-Flash-Lite}}
& Single-Prompting & NECHOv3
& \belowrule{.552} & \belowrule{.435} & \belowrule{.391}
& \underline{.752} & .254
& \belowrule{.552} & \belowrule{.411} & \belowrule{.396}
& \underline{.768} & .153 & .453 \\

\cmidrule(lr){2-14}

& \multirow{3}{*}{Multi-agent} & MedAgents
& \belowrule{.552} & \belowrule{.426} & \belowrule{.397}
& .736 & .279
& \belowrule{.532} & \belowrule{.388} & \belowrule{.372}
& \belowrule{.734} & .195 & \belowrule{.416} \\

& & MDAgents
& \belowrule{.549} & \belowrule{.457} & \belowrule{.392}
& .733 & .283
& \belowrule{.537} & \belowrule{.404} & \belowrule{.375}
& .746 & .174 & .478 \\

& & KAMAC
& \belowrule{.528} & \belowrule{.406} & \belowrule{.371}
& \belowrule{.700} & .286
& \belowrule{.514} & \belowrule{.385} & \belowrule{.349}
& \belowrule{.710} & .197 & \belowrule{.435} \\

\cmidrule(lr){2-14}

& \multirow{4}{*}{Retrieval} & Summary-RAG
& \belowrule{.522} & \belowrule{.425} & \belowrule{.362}
& \belowrule{.723} & .285
& \belowrule{.508} & \belowrule{.393} & \belowrule{.338}
& .744 & .199 & \underline{.491} \\

& & RGAR
& \belowrule{.511} & \belowrule{.413} & \belowrule{.360}
& \belowrule{.698} & .271
& \belowrule{.515} & \belowrule{.371} & \belowrule{.356}
& \belowrule{.729} & .181 & .473 \\

& & MASS-RAG
& \belowrule{.576} & \belowrule{.458} & \belowrule{.418}
& .746 & .274
& \belowrule{\textbf{.583}} & \belowrule{\textbf{.449}} & \belowrule{\textbf{.418}}
& .746 & \underline{.223} & .487 \\

& & EHR-RAG
& \underline{.584} & \textbf{.483} & \belowrule{\textbf{.424}}
& .748 & \underline{.315}
& \belowrule{.575} & \belowrule{.426} & \belowrule{.405}
& .753 & .213 & .488 \\

\cmidrule(lr){2-14}

& Memory & Traj-CoA
& \belowrule{.526} & \belowrule{.411} & \belowrule{.384}
& \belowrule{.719} & .250
& \belowrule{.525} & \belowrule{.386} & \belowrule{.369}
& .735 & .182 & .455 \\

\cmidrule(lr){2-14}

& Ours & \textbf{LONGINUS}
& \textbf{.584} & \underline{.477}
& \belowrule{\underline{.421}}
& \textbf{.773}$^{\dagger}$ & \textbf{.330}
& \belowrule{\underline{.580}}
& \belowrule{\underline{.437}}
& \belowrule{\underline{.408}}
& \textbf{.774} & \textbf{.234} & \textbf{.529} \\

\midrule

\multirow{10}{*}{\makecell{Llama-3.3\\-70B}}
& Single-Prompting & NECHOv3
& \belowrule{.533} & \belowrule{.380} & \belowrule{.383}
& \belowrule{.710} & .064
& \underline{.600} & \belowrule{.379} & \underline{.453}
& \underline{.747} & .082 & \belowrule{.406} \\

\cmidrule(lr){2-14}

& \multirow{3}{*}{Multi-agent} & MedAgents
& \belowrule{.466} & \belowrule{.303} & \belowrule{.329}
& \belowrule{.630} & .175
& \belowrule{.476} & \belowrule{.306} & \belowrule{.332}
& \belowrule{.631} & .118 & \belowrule{.378} \\

& & MDAgents
& \belowrule{.483} & \belowrule{.334} & \belowrule{.343}
& \belowrule{.654} & .178
& \belowrule{.501} & \belowrule{.320} & \belowrule{.361}
& \belowrule{.663} & .123 & \belowrule{.344} \\

& & KAMAC
& \belowrule{.500} & \belowrule{.367} & \belowrule{.358}
& \belowrule{.666} & .195
& \belowrule{.510} & \belowrule{.348} & \belowrule{.372}
& \belowrule{.678} & .122 & \belowrule{.358} \\

\cmidrule(lr){2-14}

& \multirow{4}{*}{Retrieval} & Summary-RAG
& \belowrule{.490} & \belowrule{.370} & \belowrule{.345}
& \belowrule{.674} & \underline{.247}
& \belowrule{.465} & \belowrule{.322} & \belowrule{.331}
& \belowrule{.685} & .144 & \belowrule{.436} \\

& & RGAR
& \belowrule{.501} & \belowrule{.367} & \belowrule{.381}
& \belowrule{.669} & .140
& \belowrule{.451} & \belowrule{.289} & \belowrule{.336}
& \belowrule{.608} & .034 & \belowrule{.269} \\

& & MASS-RAG
& \belowrule{\underline{.558}} & \belowrule{\underline{.433}}
& \belowrule{.416}
& \belowrule{\underline{.724}} & .202
& \belowrule{.572} & \belowrule{\underline{.422}}
& \belowrule{.411}
& \belowrule{.718} & .151 & .450 \\

& & EHR-RAG
& \belowrule{.548} & \belowrule{.424}
& \belowrule{\underline{.419}}
& \belowrule{.713} & .138
& \belowrule{.584} & \belowrule{.397} & .443
& \belowrule{.731} & .107 & \belowrule{.350} \\

\cmidrule(lr){2-14}

& Memory & Traj-CoA
& \belowrule{.494} & \belowrule{.374} & \belowrule{.359}
& \belowrule{.656} & .227
& \belowrule{.527} & \belowrule{.356} & \belowrule{.379}
& \belowrule{.686} & \underline{.184} & \underline{.453} \\

\cmidrule(lr){2-14}

& Ours & \textbf{LONGINUS}
& \textbf{.592}$^{\dagger}$
& \belowrule{\textbf{.463}$^{\dagger}$}
& \textbf{.444}$^{\dagger}$
& \textbf{.757}$^{\dagger}$
& \textbf{.262}
& \textbf{.613}
& \belowrule{\textbf{.437}}
& \textbf{.456}
& \textbf{.764}
& \textbf{.197}
& \textbf{.481} \\

\bottomrule
\end{tabular}

\caption{
Performance on MIMIC-IV by observed history depth. Formatting and $\dagger$ follow Table~\ref{tab:main_results}.
}
\vspace{-3mm}
\label{tab:visit_history}
\end{table*}

\begin{table}[t!]
\centering
\scriptsize
\setlength{\tabcolsep}{4pt}
\renewcommand{\arraystretch}{1.1}

\begin{tabular}{@{}clcccccc@{}}
\toprule
\midrule
\multirow{2}{*}{\textbf{Component}}
& \multirow{2}{*}{\textbf{Variant}}
& \textbf{Agg.}
& \multicolumn{2}{c}{\textbf{Dist.}}
& \multicolumn{3}{c}{\textbf{Temp.}} \\
\cmidrule(lr){3-3}\cmidrule(lr){4-5}\cmidrule(lr){6-8}
& & \textbf{Mic.} & \textbf{Mac.} & \textbf{Jac.}
& \textbf{Pers.} & \textbf{New} & \textbf{Rec.} \\
\hline
\midrule

\multicolumn{8}{l}{\textbf{Gemini-3.1-Flash-Lite}} \\
\midrule

Full & \textbf{LONGINUS}
& .583 & .490 & .418
& .773 & \underline{.314} & \underline{.529} \\

\cmidrule(lr){1-8}

\multirow{4}{*}{Retrieval} & w/ SQ
& .585 & \textbf{.495} & .421
& \textbf{.774} & .313 & .520 \\
& w/o Current
& .582 & .488 & .419 & .772 & .313 & .524 \\
& w/o Past-only
& .584 & .489 & .420 & .773 & \textbf{.315} & .527 \\
& w/o Unseen
& .586 & \underline{.493} & .425$^{\ddagger}$ & .773 & .306 & .523 \\

\cmidrule(lr){1-8}

\multirow{2}{*}{Memory} & w/ US
& \underline{.590}$^{\ddagger}$ & .488
& \underline{.428}$^{\ddagger}$
& \underline{.774} & .313 & \textbf{.531} \\
& w/ 4F
& .581 & .481 & .416 & .772 & .308 & .507 \\

\cmidrule(lr){1-8}

Adjudication & w/o $\mathcal{E},\mathcal{M}$
& \textbf{.597}$^{\ddagger}$ & .491
& \textbf{.432}$^{\ddagger}$
& .761$^{\dagger}$ & .305 & .522 \\

\midrule\multicolumn{8}{l}{\textbf{Llama-3.3-70B}} \\
\midrule

Full & \textbf{LONGINUS}
& .597 & \underline{.488} & .447
& .759 & \underline{.253} & \textbf{.481} \\

\cmidrule(lr){1-8}

\multirow{4}{*}{Retrieval} & w/ SQ
& .593 & .477 & .446
& .757 & .248 & .477 \\
& w/o Current
& .594 & .486 & .444
& .759 & .250 & .474 \\
& w/o Past-only
& \textbf{.598} & \textbf{.492} & .447
& \underline{.760} & \textbf{.258} & .470 \\
& w/o Unseen
& \underline{.597} & .482 & \underline{.449}
& \textbf{.760} & .250 & .471 \\

\cmidrule(lr){1-8}

\multirow{2}{*}{Memory} & w/ US
& .594 & .486 & \textbf{.451}
& .758 & .217$^{\dagger}$ & .446 \\
& w/ 4F
& .591 & .477 & .439
& .759 & .247 & \underline{.477} \\

\cmidrule(lr){1-8}

Adjudication & w/o $\mathcal{E},\mathcal{M}$
& .585$^{\dagger}$ & .460 & .439
& .751 & .189$^{\dagger}$ & .436 \\

\bottomrule
\end{tabular}

\caption{
Ablations on MIMIC-IV. w/ SQ: single query (no BRD); w/o Current/Past-only/Unseen: one history-state query removed (top-12 budget fixed); w/ US: unstructured summary; w/ 4F: card without chronic, recurrent, and acute fields; w/o $\mathcal{E},\mathcal{M}$: adjudication without evidence and memory card. Formatting follows Table~\ref{tab:main_results}; $\dagger$/$\ddagger$: full LONGINUS/the ablation is significantly better.
}
\vspace{-3mm}
\label{tab:ablation_main}
\end{table}

\subsubsection{Performance by Visit-History Depth}

Table~\ref{tab:visit_history} stratifies MIMIC-IV performance by observed history depth. With Llama-3.3-70B, LONGINUS achieves the best LLM-based score on every metric in both strata. With Gemini-3.1-Flash-Lite, it leads on all transition-specific metrics in both strata, whereas EHR-RAG leads on Macro-F1 and Jaccard in Visit~1, and MASS-RAG leads on Micro-F1, Macro-F1, and Jaccard in Visit~2+, where LVCF also exceeds LONGINUS in Micro-F1 (.586 vs.\ .580). New F1 declines from Visit~1 to Visit~2+ for all LLM-based methods except NECHOv3 with Llama-3.3-70B, consistent with the smaller share of New targets (Table~\ref{tab:preliminary_transition_distribution}). The relative strength of methods thus depends on history depth, motivating depth-stratified evaluation.

\subsection{Analysis}
\subsubsection{Ablation Studies}
\label{sec:ablation_results}

In Table~\ref{tab:ablation_main}, single-query retrieval (SQ, without BRD) lowers New and Recurrent F1 with both backbones despite slightly higher aggregate metrics with Gemini-3.1-Flash-Lite, and removing any history-state query likewise lowers Recurrent F1, though not significantly. Memory structure is backbone-dependent: an unstructured summary (US) significantly improves Micro-F1 and Jaccard with Gemini-3.1-Flash-Lite but lowers five of six metrics with Llama-3.3-70B, significantly for New F1 (.217 vs.\ .253), whereas removing history fields (4F) lowers 11 of 12 metrics, though not significantly. Removing evidence and memory from adjudication (w/o $\mathcal{E},\mathcal{M}$) lowers all temporal F1 scores with both backbones, significantly for Persistent F1 with Gemini-3.1-Flash-Lite and New F1 with Llama-3.3-70B (.189 vs.\ .253), despite significantly raising Micro-F1 and Jaccard with Gemini-3.1-Flash-Lite. Evidence-grounded adjudication is thus the most consistent contributor, while the Recurrent F1 drop across all removals suggests partially redundant evidence paths, so robustness stems from the full configuration rather than any single component.

\begin{table}[t]
\centering
\scriptsize
\setlength{\tabcolsep}{2.4pt}
\renewcommand{\arraystretch}{1.05}

\begin{tabular}{@{}cccccccc@{}}
\toprule
\midrule
\multirow{2}{*}{\textbf{Type}}
& \multirow{2}{*}{\textbf{Variant}}
& \textbf{Agg.}
& \multicolumn{2}{c}{\textbf{Dist.}}
& \multicolumn{3}{c}{\textbf{Temp.}} \\
\cmidrule(lr){3-3}\cmidrule(lr){4-5}\cmidrule(lr){6-8}
& & \textbf{Mic.} & \textbf{Mac.} & \textbf{Jac.}
& \textbf{Pers.} & \textbf{New} & \textbf{Rec.} \\
\hline
\midrule

\multicolumn{8}{l}{\textbf{Gemini-3.1-Flash-Lite}} \\
\midrule

\multirow{2}{*}{\makecell[l]{Single\\view}}
& Retrieval
& .578 & .489 & .411
& .753$^{\dagger}$ & .308 & .483$^{\dagger}$ \\

& Memory
& .556$^{\dagger}$ & .457$^{\dagger}$ & .389$^{\dagger}$
& .767 & .296$^{\dagger}$ & .529 \\

\cmidrule{1-8}

\multirow{2}{*}{\makecell[l]{No\\adjud.}}
& Union
& .552$^{\dagger}$ & .472$^{\dagger}$ & .382$^{\dagger}$
& .755$^{\dagger}$ & .300$^{\dagger}$ & .490 \\

& Intersection
& \textbf{.587} & .474 & \underline{.426}
& .765 & .303 & .523 \\

\cmidrule{1-8}

\multirow{5}{*}{Adjud.}
& Source-aware
& .582 & \textbf{.491} & .416
& .770 & \textbf{.315} & .515 \\

& Union-wide
& \underline{.587} & .483 & \textbf{.427}$^{\ddagger}$
& .771 & .304 & \textbf{.536} \\

& Memory-anch.
& .562$^{\dagger}$ & .472$^{\dagger}$ & .393$^{\dagger}$
& \underline{.773} & .302$^{\dagger}$ & .523 \\

& Retrieval-anch.
& .569$^{\dagger}$ & .486 & .400$^{\dagger}$
& .755$^{\dagger}$ & .309 & .491$^{\dagger}$ \\

\cmidrule(lr){2-8}

& \textbf{LONGINUS}
& .583 & \underline{.490} & .418
& \textbf{.773} & \underline{.314} & \underline{.529} \\

\midrule

\multicolumn{8}{l}{\textbf{Llama-3.3-70B}} \\
\midrule

\multirow{2}{*}{\makecell[l]{Single\\view}}
& Retrieval
& .587$^{\dagger}$ & .482 & .446
& .748$^{\dagger}$ & .178$^{\dagger}$ & .438 \\

& Memory
& .502$^{\dagger}$ & .389$^{\dagger}$ & .353$^{\dagger}$
& .666$^{\dagger}$ & .260 & .456 \\

\cmidrule{1-8}

\multirow{2}{*}{\makecell[l]{No\\adjud.}}
& Union
& .577$^{\dagger}$ & \underline{.490} & .419$^{\dagger}$
& .755 & \underline{.268} & .463 \\

& Intersection
& .499$^{\dagger}$ & .368$^{\dagger}$ & .370$^{\dagger}$
& .654$^{\dagger}$ & .161$^{\dagger}$ & .420 \\

\cmidrule{1-8}

\multirow{5}{*}{Adjud.}
& Source-aware
& .593 & .469 & \underline{.448}
& .753 & .223$^{\dagger}$ & .468 \\

& Union-wide
& \textbf{.599} & .487 & \textbf{.456}$^{\ddagger}$
& \textbf{.762} & .232$^{\dagger}$ & \underline{.480} \\

& Memory-anch.
& .579$^{\dagger}$ & .485 & .421$^{\dagger}$
& \underline{.759} & \textbf{.268} & .480 \\

& Retrieval-anch.
& .593 & \textbf{.491} & .439$^{\dagger}$
& .755 & .266$^{\ddagger}$ & .464 \\

\cmidrule(lr){2-8}

& \textbf{LONGINUS}
& \underline{.597} & .488 & .447
& .759 & .253 & \textbf{.481} \\

\bottomrule
\end{tabular}

\caption{
Comparison of retrieval--memory combination and adjudication strategies on MIMIC-IV.
Single view: predictions of one view alone;
Union/Intersection: set operations without adjudication;
Source-aware: LONGINUS with candidate provenance revealed;
Union-wide: adjudicates the entire union without fixing consensus;
Memory-/Retrieval-anch.: fixes one view's predictions and adjudicates only the other's additions.
\textbf{Bold}/\underline{underline} follow Table~\ref{tab:main_results}; 
$\dagger$/$\ddagger$: LONGINUS/the variant is significantly better.
}
\label{tab:comparative_main}
\vspace{-3mm}
\end{table}

\begin{table}[h!]
\centering
\scriptsize
\setlength{\tabcolsep}{4pt}
\renewcommand{\arraystretch}{1.2}

\begin{tabular}{@{}cccc@{}}
\toprule
\midrule
\textbf{\makecell{Stage}} & \textbf{\makecell{Measure (\%)}}
& \textbf{\makecell{Gemini-3.1\\-Flash-Lite}}
& \textbf{\makecell{Llama-3.3\\-70B}} \\
\hline
\midrule

\multirow{3}{*}{\makecell[l]{Candidate precision}}
& Consensus                & 54.9 & 66.7 \\
& Disagreement (before)    & 21.6 & 38.8 \\
& Disagreement (after)     & 27.8 & 49.2 \\
\midrule

\multirow{2}{*}{\makecell[l]{Within disagreement}}
& TP retained         & 64.1 & 82.2 \\
& FP rejected         & 54.3 & 46.3 \\
\midrule

\multirow{2}{*}{\makecell[l]{Final FN share}}
& Missed by both      & 83.1 & 89.3 \\
& Rejected by adjudication  & 16.9 & 10.7 \\
\midrule

\multirow{2}{*}{\makecell[l]{Final FP share}}
& Retained consensus       & 70.0 & 49.3 \\
& Accepted disagreement    & 30.0 & 50.7 \\

\bottomrule
\end{tabular}

\caption{
Stage-wise analysis of LONGINUS on MIMIC-IV using in-space labels. Disagreement (after): precision of accepted disagreements; TP retained/FP rejected: shares of correct/incorrect disagreement candidates that are accepted/rejected; final FN and FP shares are normalized by the respective final totals.
}
\label{tab:stagewise_error}
\vspace{-3mm}
\end{table}

\subsubsection{Comparative Studies}
\label{sec:comparative_studies}

Table~\ref{tab:comparative_main} compares combination strategies that share view predictions and, for adjudication variants, the adjudicator prompt and evidence. Neither view suffices alone: memory is significantly worse than LONGINUS on all aggregate and distributional metrics, and retrieval on Persistent F1. Set operations are backbone-sensitive: the intersection ties for the best Micro-F1 with Gemini-3.1-Flash-Lite but collapses with Llama-3.3-70B.

Among adjudication variants, revealing provenance lowers Recurrent F1 with both backbones and New F1 with Llama-3.3-70B (.223 vs.\ .253); Union-wide adjudication improves Jaccard but lowers New F1 with both, likely by discarding correct shared predictions; and anchoring is backbone-dependent, improving New F1 with Llama-3.3-70B but significantly lowering four metrics with Gemini-3.1-Flash-Lite. Although no variant leads every column, LONGINUS attains the best mean rank across the 12 backbone--metric columns (2.4 vs.\ 2.6 for Union-wide; $\geq$4.1 otherwise) and stays within .015 of each column's best, whereas every other variant trails by at least .035 somewhere, so it requires no prior choice of which view to trust.

\subsubsection{Stage-wise Error Analysis}

Table~\ref{tab:stagewise_error} traces predictions stage by stage on MIMIC-IV. Consensus candidates are far more precise than disagreements (54.9\% vs.\ 21.6\% with Gemini-3.1-Flash-Lite; 66.7\% vs.\ 38.8\% with Llama-3.3-70B), supporting agreement as a reliability heuristic. Adjudication raises disagreement precision to 27.8\% and 49.2\%, rejecting about half of false positives while retaining most true positives. Most final false negatives (83.1\% and 89.3\%) are missed by both views, limiting recall at candidate generation, whereas retained consensus contributes 70.0\% and 49.3\% of final false positives, bounding precision. Appendix~\ref{app:precision_recall} further quantifies the precision--recall trade-offs between the raw candidate union and final predictions, both overall and by temporal transition. We also provide case studies in Appendix~\ref{app:qualitative}.

\section{Conclusion}

Aggregate scores can obscure the multifaceted nature of next-visit diagnosis prediction and favor diagnostic persistence: last-visit carry-forward outperforms most LLM-based results in Micro-F1 without predicting new or recurrent diagnoses. Evaluating across label frequency, temporal transitions, and history depth reveals complementary retrieval and memory strengths, which LONGINUS exploits by preserving cross-view consensus and adjudicating only disagreements, achieving the best transition-specific performance among LLM-based methods in nearly all settings.

\newpage

\section*{Limitations}

\paragraph{Dataset-side Limitations.}
Mapping diagnoses to 112 CCS child labels enables consistent cross-dataset evaluation but may collapse clinically meaningful distinctions. Labels also reflect both clinical evolution and encounter-specific coding practices, which our transition categories do not fully disentangle. Due to inference costs, we evaluate on 1,000 patients per dataset, stratified by age group, sex, race, and visit depth; both datasets originate from a single medical center, which may limit generalizability. Moreover, only 214 and 256 patients have two or more observed visits, limiting statistical power for Visit~2+ and Recurrent comparisons.

\paragraph{Model-side Limitations.}
LONGINUS preserves consensus without adjudication, so erroneous shared predictions propagate to the output. Most remaining false negatives are missed by both views, revealing a candidate-generation bottleneck that adjudication alone cannot resolve. On aggregate and set-overlap metrics, LONGINUS remains below LVCF on MIMIC-III across all four backbones and only comparable to the strongest patient-internal RAG baselines elsewhere; its gains concentrate on transition-specific performance, at up to four LLM calls per patient. MIMIC-derived text may have appeared in LLM pretraining data, which could affect scores and, potentially, method comparisons.

\paragraph{Evaluation Scope.}
Our framework is model-agnostic, requiring only predicted and reference label sets, but we apply it only to rule-based and training-free LLM-based methods; supervised sequential models and fine-tuned LLMs remain future work.

\section*{Ethics Statement}

We use de-identified MIMIC-III and MIMIC-IV data under the PhysioNet Credentialed Health Data Use Agreement. Following PhysioNet’s guidance on online services, no data were sent to consumer LLM interfaces: Gemini models (Gemini-2.5-Flash-Lite and Gemini-3.1-Flash-Lite) were accessed through Google Cloud Vertex AI without data retention, whereas open-weight models (Llama-3.3-70B and GPT-OSS-120B) and \texttt{bge-small-en-v1.5} embeddings were run on local GPUs. All models were used under their respective licenses and terms of use. Appendix examples are paraphrased. LONGINUS is a research prototype not validated for clinical use.

\newpage

\bibliography{custom}

\newpage

\appendix

\section*{Appendix}
\label{sec:appendix}

\section{Extended Related Work}
\label{app:related_work}

\subsection{Next-Visit Longitudinal Diagnosis Prediction and Its Evaluation}

Longitudinal EHR models forecast future clinical events from prior patient history, including deterioration, mortality, sepsis, and future diagnoses \cite{escobar2020automated, goldstein2020development, ahmad2022alerting, adams2022prospective, hama2025enhancing, kim2026ehrnote}. Early next-visit diagnosis prediction relied on recurrent and attention-based sequence models \cite{choi2016doctor, choi2016retain, ma2017dipole}, while later work incorporated medical hierarchies, transformers, and multimodal EHR representations \cite{choi2017gram, li2020behrt, peng2021sequential, li2022integrating, koo2024next, koo2025overcoming, wu2025takecare}, as well as temporal graphs, hypergraphs, multi-view distillation, and structure-aware concept embeddings \cite{chen2024predictive, wang2025structure, tan2025boxlm, tian2026pmvgd}. 

LLMs have shown growing utility in scientific discovery and beyond \cite{liu2024drugagent, qu2025generative, zhang2025besel, reddy2025towards, low2026dual}. In clinical prediction, recent work explores LLM-based diagnosis ranking and LLM reranking over retrieved candidate outcomes
\cite{ben2024cpllm, kruse2025large, ma2025memorize,
qu2026risk, chen2026import}.

Across this literature, evaluation relies predominantly on aggregate metrics such as F1, precision/recall, ranking measures, or Jaccard \cite{hama2025enhancing, wu2025takecare, wang2026integrating}, and analyses beyond them typically examine one dimension at a time. Distributionally, prior work has examined multi-label co-occurrence \cite{sun2021interpretable, yang2025lcdl, wu2025takecare} and low-frequency diagnoses \cite{choi2017gram}. Temporally, studies have targeted newly appearing diagnoses \cite{mukherjee2023scope, ding2024large, zhang2026evicare}, including previously uncoded comorbidities \cite{rajamohan2026scaling}, but rarely distinguish Persistent diagnoses, which remain active from the current visit, from Recurrent diagnoses, which reappear after an interval of absence. History depth has likewise been analyzed separately \cite{wu2025takecare}. To our knowledge, these dimensions have not been evaluated jointly despite their clinical importance \cite{prescott2015readmission, schulz2021temporal}.

\subsection{Retrieval- and Memory-Based Approaches for Clinical Reasoning}

Retrieval-augmented generation (RAG) grounds LLMs in evidence selected at inference time \cite{lewis2020retrieval, fan2024survey, kim2025towards}. Medical RAG spans external knowledge retrieval and adaptive evidence selection \cite{xiong2024benchmarking, sohn2025rationale}. RGAR further combines factual extraction with iterative conceptual retrieval for medical question answering \cite{liang2025rgar}. Beyond medical QA, RAM-EHR, GraphCare, and KARE augment EHR-based clinical prediction with retrieved or graph-structured medical knowledge \cite{xu2024ram, jiang2024graphcare, jiang2025reasoning}, while EHR-RAG retrieves from long-horizon patient records through factual and counterfactual paths
\cite{cao2026ehr}.

Memory-based methods address LLM long-context limitations \cite{liu2024lost}. MemGPT and MemoryBank maintain long-term memory \cite{packer2023memgpt, zhong2024memorybank}, while \citet{qian2025memorag} uses memory-guided retrieval. Traj-CoA models patient trajectories through memory \cite{zeng2025trajcoa}. LONGINUS instead combines retrieval and memory as separate views with source-neutral disagreement adjudication.

\subsection{Multi-Agent Clinical Reasoning and Disagreement Adjudication}

Multi-agent debate seeks to improve LLM reasoning through iterative discussion \cite{du2023improving}. In medicine, MedAgents, MDAgents, and KAMAC coordinate specialist roles \cite{tang2024medagents, kim2024mdagents, wu2025knowledge}, while Traj-CoA and CARE-AD support longitudinal reasoning through shared memory or domain specialization \cite{zeng2025trajcoa, li2025care}. Agentic RAG further distributes evidence processing across specialized agents, as in MASS-RAG and SEMA-RAG \cite{xiao2026mass, huang2026sema}.

However, multi-agent debate does not consistently outperform simpler prompting or ensembling strategies \cite{smit2024mad}, and agreement can reinforce shared errors \cite{estornell2024multi, yang2026auditing}. Recent work therefore explores evidence-based auditing and verification \cite{yang2026auditing, valiev2026hse} and source anonymization to mitigate identity bias \cite{choi2026identity}. LONGINUS instead preserves consensus across heterogeneous evidence views and adjudicates only disagreements without revealing provenance.

\newpage

\section{Detailed Data Preprocessing}

\label{app:preprocessing}

\subsection{Raw Data Sources}

We derive the study cohorts from MIMIC-III and MIMIC-IV using only patient demographics, admission information, diagnosis codes, and discharge summaries. The corresponding source tables differ across MIMIC versions, but the same clinical information is extracted from both datasets.

\subsection{Patient Eligibility and Cohort Filtering}

Following prior preprocessing for next-visit diagnosis prediction \cite{koo2024next}, we retain adults ($\geq18$ years) with urgent, emergency, or elective-related admissions and at least two valid diagnosis admissions. Patients are excluded for invalid admission timing, missing admission, discharge-note, or diagnosis information, or the presence of pregnancy/delivery or congenital/perinatal diagnoses. The same criteria are applied to both datasets; in MIMIC-III, de-identified ages $\geq300$ are recoded to 90 to represent the masked $>89$ age group.

\subsection{Discharge Summaries}
When multiple selected discharge summaries correspond to the same admission, we order them chronologically and concatenate them into a single admission-level document. Text preprocessing is limited to regex-based cleaning; we otherwise retain the original note content, including de-identified placeholders, to avoid potential loss of clinically relevant information.

\subsection{Trajectory and Label-Space Construction}

Admissions are ordered chronologically, and consecutive visits separated by more than 180 days are treated as discontinuous following prior work \cite{weiskopf2013defining}. We retain the earliest contiguous segment, use its final admission as the next-visit target, and take up to five admissions immediately preceding the target as the observed history; earlier admissions in the segment are discarded. Table~\ref{tab:dataset_statistics} reports the resulting cohort statistics and the coverage of the retained visit history.

To establish a common diagnosis space across MIMIC-III and MIMIC-IV, we use ICD-9-CM codes directly for MIMIC-III and cross-walk MIMIC-IV ICD-10-CM codes to ICD-9-CM using the General Equivalence Mapping (GEM). Diagnoses are then mapped to the Clinical Classifications Software (CCS) taxonomy, with unmapped and \texttt{NoDx} entries removed as non-informative prediction targets. From the original 285 HCUP CCS single-level categories, we retain 112 CCS single-level labels under 14 parent categories after excluding screening, residual/unclassified, selected mental-health, injury (but not poisoning), and other out-of-scope categories. The final taxonomy was determined in consensus with physician collaborators to balance clinical relevance, diagnostic specificity, and suitability for predictive modeling (mean 8.0 children per parent; median 5.5, range 2--30). The complete label space and its empirical next-visit distribution are shown in Table~\ref{tab:label_space}.

\subsection{Sampling and Cohort Statistics}

After all eligibility, trajectory, and label-space filters, we sample 1,000 patients from each dataset using subject-level stratification by age group, sex, race, and visit depth with random seed 42. A separate patient-disjoint development cohort of 500 patients was sampled from MIMIC-III with the same procedure and used only for design selection. Final cohort and target-label statistics are summarized in Table~\ref{tab:dataset_statistics}.

\begin{table}[t]
\centering
\small
\begin{tabular}{lcc}
\toprule
\midrule
\textbf{Statistic} & \textbf{MIMIC-III} & \textbf{MIMIC-IV} \\
\hline
\midrule
Patients                     & 1,000  & 1,000  \\
Observed visit rows          & 1,308  & 1,453  \\
Mean target labels / patient & 7.745  & 6.888  \\
\bottomrule
\end{tabular}
\caption{
Final cohort statistics under a five-observed-visit cap.
}
\vspace{-3mm}
\label{tab:dataset_statistics}
\end{table}

\newpage

\clearpage
\onecolumn
\begingroup
\footnotesize
\setlength{\tabcolsep}{3pt}

\begin{longtable}{
    @{}
    p{0.16\textwidth}
    p{0.36\textwidth}
    r
    r
    r
    r
    @{}
}

\toprule
\midrule
\multirow{2}{*}{\textbf{Parent Category}} &
\multirow{2}{*}{\textbf{Child Diagnosis Label}} &
\multicolumn{2}{c}{\textbf{MIMIC-III}} &
\multicolumn{2}{c}{\textbf{MIMIC-IV}} \\
\cmidrule(lr){3-4}\cmidrule(lr){5-6}
& & \textbf{Count} & \textbf{Dist. (\%)} & \textbf{Count} & \textbf{Dist. (\%)} \\
\hline
\midrule
\endfirsthead

\multicolumn{6}{l}{\textit{Table~\thetable\ continued}} \\
\toprule
\midrule
\multirow{2}{*}{\textbf{Parent Category}} &
\multirow{2}{*}{\textbf{Child Diagnosis Label}} &
\multicolumn{2}{c}{\textbf{MIMIC-III}} &
\multicolumn{2}{c}{\textbf{MIMIC-IV}} \\
\cmidrule(lr){3-4}\cmidrule(lr){5-6}
& & \textbf{Count} & \textbf{Dist. (\%)} & \textbf{Count} & \textbf{Dist. (\%)} \\
\hline
\midrule
\endhead

\midrule
\multicolumn{6}{r}{\textit{Continued on next page}} \\
\endfoot

\bottomrule
\endlastfoot

\multicolumn{6}{l}{\textit{Diseases of the blood and blood-forming organs}}\\
\addlinespace[1pt]

& Acute post-hemorrhagic anemia
& 94 & 1.21 & 63 & 0.91 \\

& Anemia
& 301 & 3.89 & 292 & 4.24 \\

& Coagulation and hemorrhagic disorders
& 137 & 1.77 & 126 & 1.83 \\

& White blood cell disorders
& 49 & 0.63 & 87 & 1.26 \\

\addlinespace

\cmidrule(lr){1-6}
\multicolumn{6}{l}{\textit{Diseases of the circulatory system}}\\
\addlinespace[1pt]

& Acute cerebrovascular disease
& 46 & 0.59 & 15 & 0.22 \\

& Acute myocardial infarction
& 71 & 0.92 & 58 & 0.84 \\

& Aneurysm
& 35 & 0.45 & 23 & 0.33 \\

& Arterial embolism
& 9 & 0.12 & 4 & 0.06 \\

& Cardiac arrest
& 12 & 0.15 & 1 & 0.01 \\

& Cardiac dysrhythmia
& 389 & 5.02 & 304 & 4.41 \\

& Carditis
& 100 & 1.29 & 52 & 0.75 \\

& Conduction disorders
& 87 & 1.12 & 93 & 1.35 \\

& Congestive heart failure; non-hypertensive
& 375 & 4.84 & 253 & 3.67 \\

& Coronary atherosclerosis
& 341 & 4.40 & 318 & 4.62 \\

& Heart valve disease
& 147 & 1.90 & 148 & 2.15 \\

& Hypertension
& 389 & 5.02 & 391 & 5.68 \\

& Hypertension complications
& 188 & 2.43 & 216 & 3.14 \\

& Late effects of cerebrovascular disease
& 41 & 0.53 & 27 & 0.39 \\

& Peripheral atherosclerosis
& 81 & 1.05 & 92 & 1.34 \\

& Phlebitis
& 107 & 1.38 & 107 & 1.55 \\

& Pre-cerebral occlusion
& 15 & 0.19 & 20 & 0.29 \\

& Pulmonary heart disease
& 84 & 1.08 & 83 & 1.20 \\

& Transient ischemic attack
& 6 & 0.08 & 4 & 0.06 \\

\addlinespace

\cmidrule(lr){1-6}
\multicolumn{6}{l}{\textit{Diseases of the digestive system}}\\
\addlinespace[1pt]

& Abdominal hernia
& 21 & 0.27 & 25 & 0.36 \\

& Appendicitis
& 1 & 0.01 & 2 & 0.03 \\

& Biliary disease
& 52 & 0.67 & 34 & 0.49 \\

& Diverticulosis
& 35 & 0.45 & 26 & 0.38 \\

& Gastric and duodenal ulcer
& 19 & 0.25 & 17 & 0.25 \\

& Gastroenteritis
& 6 & 0.08 & 11 & 0.16 \\

& Gastrointestinal hemorrhage
& 93 & 1.20 & 45 & 0.65 \\

& Intestinal infection
& 64 & 0.83 & 38 & 0.55 \\

& Intestinal obstruction
& 34 & 0.44 & 28 & 0.41 \\

& Pancreas disease
& 25 & 0.32 & 23 & 0.33 \\

& Peritonitis
& 31 & 0.40 & 24 & 0.35 \\

& Ulcerative colitis
& 21 & 0.27 & 15 & 0.22 \\

\addlinespace

\cmidrule(lr){1-6}
\multicolumn{6}{l}{\textit{Diseases of the genitourinary system}}\\
\addlinespace[1pt]

& Acute renal failure
& 257 & 3.32 & 225 & 3.27 \\

& Chronic kidney disease
& 195 & 2.52 & 231 & 3.35 \\

& Nephritis
& 21 & 0.27 & 31 & 0.45 \\

& Urinary stone
& 12 & 0.15 & 13 & 0.19 \\

& Urinary tract infection
& 193 & 2.49 & 107 & 1.55 \\

\addlinespace

\cmidrule(lr){1-6}
\multicolumn{6}{l}{
\textit{Diseases of the musculoskeletal system and connective tissue}
}\\
\addlinespace[1pt]

& Infectious arthritis
& 25 & 0.32 & 38 & 0.55 \\

& Pathological fracture
& 11 & 0.14 & 9 & 0.13 \\

& Rheumatoid arthritis
& 23 & 0.30 & 19 & 0.28 \\

& Systemic lupus erythematosus
& 8 & 0.10 & 10 & 0.15 \\

\addlinespace

\cmidrule(lr){1-6}
\multicolumn{6}{l}{
\textit{Diseases of the nervous system and sense organs}
}\\
\addlinespace[1pt]

& Coma and brain damage
& 20 & 0.26 & 11 & 0.16 \\

& Encephalitis
& 2 & 0.03 & 4 & 0.06 \\

& Epilepsy and convulsive disorders
& 92 & 1.19 & 51 & 0.74 \\

& Meningitis
& 3 & 0.04 & 2 & 0.03 \\

& Multiple sclerosis
& 9 & 0.12 & 5 & 0.07 \\

& Paralysis
& 22 & 0.28 & 14 & 0.20 \\

& Parkinson's disease
& 11 & 0.14 & 9 & 0.13 \\

\addlinespace

\cmidrule(lr){1-6}
\multicolumn{6}{l}{\textit{Diseases of the respiratory system}}\\
\addlinespace[1pt]

& Adult respiratory failure
& 254 & 3.28 & 93 & 1.35 \\

& Aspiration pneumonia
& 88 & 1.14 & 15 & 0.22 \\

& Asthma
& 80 & 1.03 & 76 & 1.10 \\

& Chronic obstructive pulmonary disease
& 195 & 2.52 & 141 & 2.05 \\

& Lung diseases due to external agents
& 2 & 0.03 & 6 & 0.09 \\

& Pleurisy
& 138 & 1.78 & 73 & 1.06 \\

& Pneumonia
& 160 & 2.07 & 73 & 1.06 \\

\addlinespace

\cmidrule(lr){1-6}
\multicolumn{6}{l}{
\textit{Diseases of the skin and subcutaneous tissue}
}\\
\addlinespace[1pt]

& Skin infection
& 43 & 0.56 & 39 & 0.57 \\

& Skin ulcer
& 110 & 1.42 & 84 & 1.22 \\

\addlinespace

\cmidrule(lr){1-6}
\multicolumn{6}{l}{
\textit{Endocrine; nutritional; and metabolic diseases and immunity disorders}
}\\
\addlinespace[1pt]

& Diabetes mellitus with complications
& 112 & 1.45 & 141 & 2.05 \\

& Diabetes mellitus without complications
& 231 & 2.98 & 185 & 2.69 \\

& Fluid and electrolyte disorders
& 352 & 4.54 & 279 & 4.05 \\

& Gout and other crystal arthropathies
& 52 & 0.67 & 65 & 0.94 \\

& Hyperlipidemia
& 275 & 3.55 & 380 & 5.52 \\

& Immunity disorders
& 3 & 0.04 & 8 & 0.12 \\

& Nutritional deficiencies
& 61 & 0.79 & 74 & 1.07 \\

& Thyroid disorders
& 131 & 1.69 & 142 & 2.06 \\

\addlinespace

\cmidrule(lr){1-6}
\multicolumn{6}{l}{\textit{Infectious and parasitic diseases}}\\
\addlinespace[1pt]

& HIV infection
& 21 & 0.27 & 13 & 0.19 \\

& Hepatitis
& 79 & 1.02 & 63 & 0.91 \\

& Mycoses
& 55 & 0.71 & 41 & 0.60 \\

& Septicemia
& 215 & 2.78 & 85 & 1.23 \\

& Tuberculosis
& 6 & 0.08 & 3 & 0.04 \\

& Viral infection
& 14 & 0.18 & 23 & 0.33 \\

\addlinespace

\cmidrule(lr){1-6}
\multicolumn{6}{l}{\textit{Injury and poisoning}}\\
\addlinespace[1pt]

& Poisoning by non-medications
& 1 & 0.01 & 0 & 0.00 \\

& Poisoning by other medications
& 9 & 0.12 & 15 & 0.22 \\

& Poisoning by psychiatric medications
& 3 & 0.04 & 0 & 0.00 \\

\addlinespace

\cmidrule(lr){1-6}
\multicolumn{6}{l}{\textit{Mental illness}}\\
\addlinespace[1pt]

& Alcohol-related disorders
& 59 & 0.76 & 92 & 1.34 \\

& Delirium/dementia/amnesia/cognitive disorders
& 72 & 0.93 & 58 & 0.84 \\

& Substance-related disorders
& 42 & 0.54 & 40 & 0.58 \\

\addlinespace

\cmidrule(lr){1-6}
\multicolumn{6}{l}{\textit{Neoplasms}}\\
\addlinespace[1pt]

& Bladder cancer
& 11 & 0.14 & 12 & 0.17 \\

& Bone and connective tissue cancer
& 2 & 0.03 & 10 & 0.15 \\

& Brain/nervous system cancer
& 13 & 0.17 & 2 & 0.03 \\

& Breast cancer
& 27 & 0.35 & 35 & 0.51 \\

& Bronchus/lung cancer
& 44 & 0.57 & 22 & 0.32 \\

& Cervical cancer
& 3 & 0.04 & 5 & 0.07 \\

& Colon cancer
& 25 & 0.32 & 19 & 0.28 \\

& Esophageal cancer
& 13 & 0.17 & 5 & 0.07 \\

& Female genital cancer
& 1 & 0.01 & 1 & 0.01 \\

& Gastrointestinal and peritoneal cancer
& 7 & 0.09 & 8 & 0.12 \\

& Head and neck cancer
& 6 & 0.08 & 8 & 0.12 \\

& Hodgkin's disease
& 0 & 0.00 & 5 & 0.07 \\

& Kidney and renal cancer
& 14 & 0.18 & 53 & 0.77 \\

& Leukemias
& 15 & 0.19 & 55 & 0.80 \\

& Liver or inflammatory bowel disease cancer
& 12 & 0.15 & 17 & 0.25 \\

& Male genital cancer
& 1 & 0.01 & 1 & 0.01 \\

& Multiple myeloma
& 4 & 0.05 & 31 & 0.45 \\

& Non-Hodgkin lymphoma
& 18 & 0.23 & 122 & 1.77 \\

& Non-epithelial cancer
& 9 & 0.12 & 34 & 0.49 \\

& Ovary cancer
& 5 & 0.06 & 6 & 0.09 \\

& Pancreatic cancer
& 3 & 0.04 & 1 & 0.01 \\

& Prostate cancer
& 35 & 0.45 & 28 & 0.41 \\

& Rectal or anal cancer
& 7 & 0.09 & 11 & 0.16 \\

& Secondary malignancy
& 75 & 0.97 & 122 & 1.77 \\

& Skin melanoma
& 10 & 0.13 & 42 & 0.61 \\

& Stomach cancer
& 3 & 0.04 & 4 & 0.06 \\

& Testicular cancer
& 0 & 0.00 & 2 & 0.03 \\

& Thyroid cancer
& 6 & 0.08 & 4 & 0.06 \\

& Urinary organ cancer
& 1 & 0.01 & 4 & 0.06 \\

& Uterine cancer
& 5 & 0.06 & 8 & 0.12 \\

\addlinespace

\cmidrule(lr){1-6}
\multicolumn{6}{l}{
\textit{Symptoms; signs; and ill-defined conditions and factors influencing health status}
}\\

& Gangrene
& 18 & 0.23 & 24 & 0.35 \\

& Shock
& 79 & 1.02 & 36 & 0.52 \\

\bottomrule

\caption{
Complete diagnosis label space and distribution of next-visit ground-truth diagnoses in MIMIC-III and MIMIC-IV. Count denotes the number of next-visit targets containing each diagnosis,
and Dist. denotes its proportion among all positive next-visit diagnosis assignments in the corresponding dataset.
}
\label{tab:label_space}\\

\end{longtable}
\endgroup

\clearpage
\newpage
\twocolumn

\section{Evaluation Metrics}
\label{app:eval_metrics}

For patient $i=1,\ldots,n$, let $Y_i=\mathcal{S}_{T+1}$ and $\hat{Y}_i=\hat{\mathcal{S}}_{T+1}\subseteq\mathcal{L}$ denote the reference and predicted next-visit diagnosis sets, with out-of-space predictions removed and $|\mathcal{L}|=112$.

\paragraph{Micro-F1.}
Counts are pooled over patients:
\begin{equation}
\mathrm{F1}_{\mathrm{micro}}=\frac{2\sum_i\mathrm{TP}_i}{\sum_i\left(2\mathrm{TP}_i+\mathrm{FP}_i+\mathrm{FN}_i\right)},
\label{eq:microf1}
\end{equation}
where $\mathrm{TP}_i=|\hat{Y}_i\cap Y_i|$, $\mathrm{FP}_i=|\hat{Y}_i\setminus Y_i|$, and $\mathrm{FN}_i=|Y_i\setminus\hat{Y}_i|$.

\paragraph{Macro-F1.}
For each $\ell\in\mathcal{L}$, $\mathrm{TP}_\ell$, $\mathrm{FP}_\ell$, and $\mathrm{FN}_\ell$ count patients with $\ell\in\hat{Y}_i\cap Y_i$, $\ell\in\hat{Y}_i\setminus Y_i$, and $\ell\in Y_i\setminus\hat{Y}_i$, respectively:
\begin{equation}
\mathrm{F1}_{\mathrm{macro}}=\frac{1}{|\mathcal{L}|}\sum_{\ell\in\mathcal{L}}\frac{2\mathrm{TP}_\ell}{2\mathrm{TP}_\ell+\mathrm{FP}_\ell+\mathrm{FN}_\ell},
\end{equation}
where terms with a zero denominator are set to 0.

\paragraph{Jaccard.}
\begin{equation}
\mathrm{J}=\frac{1}{n}\sum_{i=1}^{n}\frac{|\hat{Y}_i\cap Y_i|}{|\hat{Y}_i\cup Y_i|},
\end{equation}
where a patient with $\hat{Y}_i\cup Y_i=\emptyset$ contributes 0.

\paragraph{Transition-specific F1.}
Using the history-state partition of Eqs.~(5)--(7) for each patient, let $\mathcal{L}^{\mathrm{pers}}_i=\mathcal{L}_{\mathrm{cur}}=\mathcal{S}_T$,
$\mathcal{L}^{\mathrm{new}}_i=\mathcal{L}_{\mathrm{uns}}=\mathcal{L}\setminus\mathcal{H}_T$, and
$\mathcal{L}^{\mathrm{rec}}_i=\mathcal{L}_{\mathrm{past}}=\mathcal{H}_{T-1}\setminus\mathcal{S}_T$.
For $\tau\in\{\mathrm{pers},\mathrm{new},\mathrm{rec}\}$, $\mathrm{F1}^{\tau}$ applies Eq.~\eqref{eq:microf1} after replacing $Y_i$ and $\hat{Y}_i$ with $Y_i\cap\mathcal{L}^{\tau}_i$ and $\hat{Y}_i\cap\mathcal{L}^{\tau}_i$, so false positives are also attributed to a transition. New is defined operationally by the observed coding history and thus covers both newly emerging and previously uncoded diagnoses.

\paragraph{History depth.}
All metrics are recomputed within Visit~1 ($T=1$) and Visit~2+ ($T\ge2$). In Visit~1, $\mathcal{H}_T=\mathcal{S}_T$, so $\mathcal{L}_i^{\mathrm{rec}} = \emptyset$ and Recurrent F1 is reported as --.

\newpage

\section{Baselines}
\label{app:baselines}

We evaluate all methods on the same observed patient histories $V_{1:T}$. Rule-based heuristics use only documented diagnosis codes, whereas LLM-based methods additionally use demographics and clinical notes. We make only minimal task-specific modifications to adapt the LLM-based baselines to our task.

\subsection{Rule-Based Heuristics}
We consider two deterministic baselines using only observed diagnosis codes:
\begin{itemize}
    \item Last-Visit Carry-Forward (LVCF)
    predicts diagnoses from the most recent visit: \\
    $
    \hat{S}_{T+1}^{\mathrm{LVCF}} = S_T.
    $
    \item Observed-History Union (OHU)
    predicts all previously recorded diagnoses: \\
    $
    \hat{S}_{T+1}^{\mathrm{OHU}}
    = H_T = \bigcup_{t=1}^{T} S_t.
    $
\end{itemize}

\subsection{Single Prompting}
\begin{itemize}[leftmargin=*, noitemsep, topsep=0pt]
\item NECHOv3 \cite{koo2025towards} employs a hierarchical prompting strategy that reasons through the CCS hierarchy, moving from CCS multi-level parent systems to CCS single-level child labels, while explicitly considering longitudinal diagnosis status.
\end{itemize}

\subsection{Multi-Agent Systems}
Specialist roles are organized around CCS multi-level parent diagnostic systems, while final predictions remain constrained to the CCS single-level child labels.
\begin{itemize}[leftmargin=*, noitemsep, topsep=0pt]
\item MedAgents \cite{tang2024medagents} recruits medical specialists that independently propose diagnoses, followed by synthesis, consensus/revision, and a final reviewer.
\item MDAgents \cite{kim2024mdagents} adapts the collaboration strategy according to estimated task complexity. Basic cases use direct reasoning, whereas more complex cases employ specialist discussion and moderator synthesis.
\item KAMAC \cite{wu2025knowledge} dynamically recruits relevant medical specialists, aggregates their analyses, and iteratively refines them through coordinator synthesis, critique, and final review.
\end{itemize}

\subsection{Retrieval}

All RAG baselines retrieve the top 12 chunks with \texttt{bge-small-en-v1.5}\footnote{Available at \url{https://huggingface.co/BAAI/bge-small-en-v1.5}.} \cite{xiao2024c}. EHR-RAG and MASS-RAG retrieve from the patient's own observed EHR trajectory, whereas Summary-RAG and RGAR retrieve from the frozen Textbooks corpus of MedRAG \cite{xiong2024benchmarking}, comprising 18 medical textbooks from MedQA \cite{jin2021disease}.

\begin{itemize}[leftmargin=*, noitemsep, topsep=0pt]
\item Summary-RAG summarizes the full patient record, uses the summary as a query to retrieve from the Textbooks corpus, and predicts next-visit diagnoses from the summary and retrieved passages, following the medical QA setting of \citet{xiong2024benchmarking}.
\item RGAR \cite{liang2025rgar} iteratively combines factual extraction from the patient record, conceptual retrieval from the Textbooks corpus, and query regeneration before producing the final prediction.
\item MASS-RAG \cite{xiao2026mass} employs a multi-agent system whose summarization, extraction, and reasoning agents we adapt to next-visit prediction. From the same retrieved patient evidence, they construct, in parallel, (1) a temporal summary, (2) explicit supporting and counter-evidence, and (3) diagnostic candidates, which a final synthesizer integrates into the prediction.
\item EHR-RAG \cite{cao2026ehr} uses visit-aware patient chunks and two retrieval paths: a factual path and a counterfactual path, each retrieving top-6 evidence chunks. It additionally applies temporal reranking to favor recent evidence.
\end{itemize}

\subsection{Memory}
\begin{itemize}[leftmargin=*, noitemsep, topsep=0pt]
\item Traj-CoA \cite{zeng2025trajcoa} processes longitudinal EHRs in a visit-by-visit manner using worker agents, progressively updating a compact trajectory summary and shared long-term memory. A final manager agent synthesizes the accumulated summary and memory for prediction. 
\end{itemize}

\subsection{LLM Calls}
Table~\ref{tab:llm_calls} summarizes end-to-end LLM calls per patient, accounting for observed visits, specialist agents, and iterative rounds.

\begin{table}[h]
\centering
\scriptsize
\setlength{\tabcolsep}{4pt}
\renewcommand{\arraystretch}{1.05}
\begin{tabular}{@{}ccl@{}}
\toprule
\midrule
\textbf{Type} & \textbf{Method} & \textbf{LLM calls per patient} \\
\hline
\midrule
\multirow{1}{*}{Single-Prompting}
  & NECHOv3 & $1$ \\
\midrule
\multirow{3}{*}{Multi-agent}
  & MedAgents & $3+E+\sum_{r=1}^{R} B_r$ \\
  & MDAgents  & $3+E+\sum_{r=1}^{R} B_r$ \\
  & KAMAC     & $4+E+\sum_{r=1}^{R} B_r$ \\
\midrule
\multirow{4}{*}{Retrieval}
  & Summary-RAG & $2$ \\
  & RGAR        & $1+C_{\mathrm{extract}}+C_{\mathrm{query}}$ \\
  & MASS-RAG    & $4$ \\
  & EHR-RAG     & $1$ \\
\midrule
Memory & Traj-CoA & $N+1$ \\
\midrule
\multirow{2}{*}{Ablation}
  & Retrieval-view only & $1$ \\
  & Memory-view only    & $2$ \\
\midrule
Ours & LONGINUS & $3+\mathbb{I}[|\mathcal{D}|>0]$ \\
\bottomrule
\end{tabular}
\caption{
End-to-end LLM calls per patient. $N$: number of observed visits; $E$: number of initial specialist calls; $R$: number of executed rounds; $B_r$: number of calls in round $r$; $C_{\mathrm{extract}}$, $C_{\mathrm{query}}$: RGAR extraction and query-generation calls; $\mathcal{D}$: diagnoses on which the two views disagree; $\mathbb{I}[\cdot]$: indicator function. Parallel calls are counted separately; embedding calls are excluded.
}
\label{tab:llm_calls}
\end{table}

\subsection{Inference Token Usage}
\label{app:token_usage}

Table~\ref{tab:token_usage} reports mean input/output token usage per patient for LONGINUS and selected baselines, computed from recorded API usage over all 1,000 patients; for retrieval--memory methods, all prediction, memory, and adjudication calls are included, and embedding and indexing costs are excluded. In total (input + output) tokens, LONGINUS uses 54--76\% fewer tokens than the multi-agent baselines (KAMAC, MedAgents, and MDAgents) and is comparable to MASS-RAG (22\% fewer to 13\% more), while generating fewer output tokens than MASS-RAG in every setting. It uses 13--41\% more tokens than Traj-CoA and 1.7--3.5\% fewer than Union-wide, since consensus labels are not re-adjudicated.

\begin{table}[h!]
\centering
\scriptsize
\setlength{\tabcolsep}{3.5pt}
\renewcommand{\arraystretch}{1.08}

\begin{tabular}{@{}ccccccccc@{}}
\toprule
\midrule
\multirow{2}{*}{\textbf{Method}}
& \multicolumn{2}{c}{\textbf{\makecell{Gemini-2.5\\-Flash-Lite}}}
& \multicolumn{2}{c}{\textbf{\makecell{Gemini-3.1\\-Flash-Lite}}}
& \multicolumn{2}{c}{\textbf{\makecell{GPT-OSS\\-120B}}}
& \multicolumn{2}{c}{\textbf{\makecell{Llama-3.3\\-70B}}} \\
\cmidrule(lr){2-3}
\cmidrule(lr){4-5}
\cmidrule(lr){6-7}
\cmidrule(lr){8-9}
& \textbf{In} & \textbf{Out}
& \textbf{In} & \textbf{Out}
& \textbf{In} & \textbf{Out}
& \textbf{In} & \textbf{Out} \\
\hline
\midrule

\multicolumn{9}{l}{\textbf{MIMIC-III}} \\
\midrule

KAMAC
& 49.87 & 3.35
& 50.35 & 1.52
& 61.32 & 9.22
& 62.83 & 1.46 \\

MedAgents
& 40.47 & 3.95
& 45.37 & 1.48
& 46.72 & 5.74
& 44.96 & 1.75 \\

MDAgents
& 65.97 & 4.09
& 71.38 & 1.18
& 67.44 & 5.10
& 73.68 & 0.55 \\

Traj-CoA
& 14.33 & 2.02
& 13.38 & 0.90
& 12.39 & 3.53
& 12.76 & 0.83 \\

MASS-RAG
& 18.60 & 5.40
& 14.79 & 1.50
& 14.42 & 7.31
& 14.20 & 1.56 \\

\cmidrule(lr){1-9}

Union-wide
& 18.50 & 1.19
& 18.19 & 0.77
& 17.85 & 5.42
& 17.18 & 0.64 \\

\textbf{LONGINUS}
& 18.20 & 1.01
& 17.81 & 0.61
& 17.50 & 4.97
& 16.95 & 0.55 \\

\midrule
\multicolumn{9}{l}{\textbf{MIMIC-IV}} \\
\midrule

KAMAC
& 45.69 & 3.04
& 46.07 & 1.45
& 54.25 & 8.06
& 58.10 & 1.36 \\

MedAgents
& 36.02 & 3.55
& 40.25 & 1.36
& 42.04 & 5.42
& 41.11 & 1.60 \\

MDAgents
& 59.45 & 3.78
& 53.66 & 0.93
& 60.54 & 4.78
& 65.56 & 0.51 \\

Traj-CoA
& 14.04 & 2.11
& 12.98 & 0.92
& 12.97 & 3.81
& 12.59 & 0.86 \\

MASS-RAG
& 18.09 & 5.26
& 14.32 & 1.46
& 14.76 & 7.11
& 14.23 & 1.49 \\

\cmidrule(lr){1-9}

Union-wide
& 17.68 & 1.07
& 17.52 & 0.72
& 17.46 & 5.03
& 16.85 & 0.60 \\

\textbf{LONGINUS}
& 17.40 & 0.91
& 17.15 & 0.58
& 17.08 & 4.62
& 16.63 & 0.53 \\

\bottomrule
\end{tabular}

\caption{
Mean recorded input/output LLM tokens per patient (thousands). Embedding and indexing costs are excluded.
}
\label{tab:token_usage}
\vspace{-3mm}
\end{table}

\newpage

\section{LONGINUS Prompt Templates}
\label{app:prompts}

The transition-aware criteria in the adjudication prompt are conditioned on each candidate’s history state (Current, Past-only, or Unseen; Section~\ref{sec:methodology_retrieval_view}), which is observable at prediction time; no next-visit information is provided to any component.

\noindent
{\small
\begin{promptbox}[History-State-Specific Retrieval Query Templates]

\textbf{Current.}
Recent patient-specific evidence of ongoing activity, continued treatment
or monitoring, unresolved disease, improvement, resolution, completed
treatment, or discontinuation for these currently coded diagnoses:

\texttt{\{\{CURRENT\_LABELS\}\}}

\vspace{0.7em}
\textbf{Past-only.}
Patient-specific evidence from prior episodes and recent evidence of
recurrence, reactivation, exacerbation, related treatment, or continued
absence for these previously coded diagnoses:

\texttt{\{\{PAST\_ONLY\_LABELS\}\}}

\vspace{0.7em}
\textbf{Unseen.}
Recent patient-specific evidence of newly emerging or previously uncoded
disease, including new or worsening symptoms, abnormal tests, active
workup, new treatment, organ complications, latent comorbidities, or
natural downstream conditions.

\end{promptbox}
}
\normalsize

\noindent
{\small
\begin{promptbox}[Patient-Internal Retrieval-View Prediction Prompt]

\textbf{System.}
You are an experienced critical care physician working in an Intensive Care Unit (ICU) predicting diagnoses likely to be actively coded at a patient's next hospital visit. Favor clinically supported recall and return only valid JSON.

\vspace{0.5em}
\textbf{Task.}
Predict the child diagnosis labels most likely to be actively documented and coded at the patient's next hospital visit.

\vspace{0.5em}
\textbf{Instructions.}
\begin{itemize}[leftmargin=*, itemsep=1pt, topsep=2pt]
    \item You are given patient-note evidence retrieved through complementary temporal searches. Interpret all evidence chronologically.
    \item Include clinically plausible persistent conditions, unresolved complications, recurrent acute events, and newly relevant diagnoses when supported by the patient-specific record.
    \item Do not require one explicit diagnostic sentence when several consistent findings, treatments, tests, or longitudinal changes support the label.
    \item Actively consider comorbidity: a disease may be newly coded because it was previously latent or uncoded, emerges as a downstream condition, or is a distinct complication of an established disease.
    \item General medical plausibility alone is insufficient; patient-specific compatible evidence is required.
    \item Use only exact labels from the fixed label space.
    \item Exclude clearly negated, ruled-out, resolved, completed, or contradicted conditions.
    \item Return multiple labels when supported; do not impose a fixed prediction count.
    \item Do not output evidence, rationale, confidence, or commentary.
\end{itemize}

\vspace{0.5em}
\textbf{Inputs.}

\texttt{\{\{TRANSITION\_EVIDENCE\_JSON\}\}}

\texttt{\{\{LABEL\_SPACE\_JSON\}\}}

\vspace{0.5em}
\textbf{Output.}
\begin{verbatim}
{
  "predictions": [
    {"label": "exact child label"}
  ]
}
\end{verbatim}

\end{promptbox}
}
\normalsize

\noindent
{\small
\begin{promptbox}[Longitudinal Memory Construction Prompt]

\textbf{System.}
You are an experienced critical care physician working in an Intensive Care Unit (ICU) who extracts patient-specific longitudinal clinical memory from observed hospital visits. Return only valid JSON.

\vspace{0.5em}
\textbf{Task.}
Given a longitudinal clinical record and a fixed diagnosis label space, construct a patient-specific memory card that preserves clinically supported signals most relevant to next-visit diagnosis prediction.

\vspace{0.5em}
\textbf{Instructions.}
\begin{itemize}[leftmargin=*, itemsep=1pt, topsep=2pt]
    \item Treat earlier visits as longitudinal history.
    \item Capture clinically plausible chronic, recurrent, acute, unresolved, treated, and monitored information when it has patient-specific support.
    \item Include tentative signals only when they are patient-specific and clinically relevant; avoid weakly supported, generic, or speculative associations.
    \item Do not discard remote chronic history solely because it is not repeated in the most recent visit.
    \item Do not predict the next visit.
    \item Do not add generic textbook associations or information unsupported by the record.
    \item Keep each item concise.
    \item Every schema field must be an array of concise strings, not objects.
\end{itemize}

\vspace{0.5em}
\textbf{Inputs.}

Longitudinal clinical record:

\texttt{\{\{CLINICAL\_NOTE\}\}}

Fixed diagnosis label space:

\texttt{\{\{LABEL\_SPACE\_JSON\}\}}

\vspace{0.5em}
\textbf{Output.}
\begin{verbatim}
{
  "chronic_conditions": [],
  "recurrent_events": [],
  "acute_events": [],
  "abnormal_trends": [],
  "prior_treatments_or_procedures": [],
  "unresolved_problems": [],
  "risk_factors": []
}
\end{verbatim}

\end{promptbox}
}
\normalsize

\noindent
{\small
\begin{promptbox}[Longitudinal Memory-View Prediction Prompt]

\textbf{System.}
You are an experienced critical care physician working in an Intensive Care Unit (ICU) predicting which diagnoses are most likely to be actively coded at the patient's next hospital visit. Return only valid JSON.

\vspace{0.5em}
\textbf{Task.}
Predict all clinically supported next-visit diagnoses from the fixed diagnosis label space using the structured patient memory.

\vspace{0.5em}
\textbf{Instructions.}
\begin{itemize}[leftmargin=*, itemsep=1pt, topsep=2pt]
    \item Do not impose a fixed prediction count.
    \item Every predicted diagnosis must be copied exactly from the fixed label space, which is treated as a closed vocabulary.
    \item Do not output raw note phrases, abbreviations, ICD names, synonyms, subtypes, or more specific disease names unless copied exactly from the label space.
    \item If a clinical concept is present but the exact label is absent, choose the closest exact label only when clearly supported; otherwise omit it.
    \item Prioritize diagnoses likely to appear at the next visit rather than merely historical diagnoses.
    \item Include multiple diagnoses when clinically plausible because of persistent comorbidities, unresolved complications, recurrent acute events, or discharge-related risks.
    \item Use longitudinal patterns, unresolved problems, recurrent conditions, and recent acute events.
    \item Avoid weakly supported or speculative diagnoses.
    \item Do not output hidden chain-of-thought.
\end{itemize}

\vspace{0.5em}
\textbf{Inputs.}

Patient memory:

\texttt{\{\{PATIENT\_MEMORY\_JSON\}\}}

Fixed diagnosis label space:

\texttt{\{\{LABEL\_SPACE\_JSON\}\}}

\vspace{0.5em}
\textbf{Output.}
\begin{verbatim}
{
  "next_visit_diagnoses": []
}
\end{verbatim}

\end{promptbox}
}
\normalsize

\noindent
{\small
\begin{promptbox}[Source-Neutral Disagreement Adjudication Prompt]

\textbf{System.}
You are an experienced critical care physician adjudicating uncertain next-visit diagnosis-code candidates. Judge each candidate independently from patient evidence and return only valid JSON.

\vspace{0.5em}
\textbf{Task.}
Decide whether every disagreement candidate should be included or excluded from the final next-visit diagnosis set. Candidate origin is intentionally hidden. Apply the same standard to every candidate; list membership is not evidence. This is a coding-transition prediction, not a broad differential diagnosis.

\vspace{0.5em}
\textbf{Transition-Aware Criteria.}
\begin{itemize}[leftmargin=*, itemsep=1pt, topsep=2pt]
    \item \textbf{Current:} Include when the observed course supports persistence, active treatment, monitoring, unresolved activity, or likely repeat coding. Exclude when there is credible resolution, improvement with discontinuation, or non-continuation evidence.
    
    \item \textbf{Past-only:} Do not exclude solely because the condition is absent from the latest visit. Include when prior repeated coding, an episodic or recurrence-prone course, persistent susceptibility, related treatment, or clinically connected activity makes repeat coding plausible. Exclude when the condition is remote and isolated or there is credible resolution, treatment completion, or sustained contradictory evidence.
    
    \item \textbf{Unseen:} Include for a patient-specific emerging condition, latent or previously uncoded comorbidity, downstream complication, or distinct new event. Generic medical plausibility alone is insufficient.
\end{itemize}

\vspace{0.5em}
\textbf{Additional Rules.}
\begin{itemize}[leftmargin=*, itemsep=1pt, topsep=2pt]
    \item Consensus labels are fixed and are not candidates.
    \item Absence from either evidence source is not by itself evidence for exclusion.
    \item Do not invent or rename labels.
    \item Return exactly one decision for every disagreement candidate.
    \item Use only \texttt{include} or \texttt{exclude}.
    \item Do not provide rationale, evidence, confidence, or commentary.
\end{itemize}

\vspace{0.5em}
\textbf{Inputs.}

Observed trajectory metadata:

\texttt{\{\{TRANSITION\_METADATA\_JSON\}\}}

Fixed consensus labels:

\texttt{\{\{CONSENSUS\_LABELS\_JSON\}\}}

Patient-local retrieved evidence:

\texttt{\{\{RETRIEVED\_EVIDENCE\_JSON\}\}}

Longitudinal memory card:

\texttt{\{\{PATIENT\_MEMORY\_JSON\}\}}

Source-neutral disagreement candidates:

\texttt{\{\{DISAGREEMENT\_CANDIDATES\_JSON\}\}}

\vspace{0.5em}
\textbf{Output.}
\begin{verbatim}
{
  "candidate_decisions": [
    {
      "label": "exact candidate label",
      "decision": "include | exclude"
    }
  ]
}
\end{verbatim}

\end{promptbox}
}
\normalsize


\section{Full Results}
\label{app:appendix_full_results}

Tables~\ref{tab:appendix_main_results}--\ref{tab:freq_macro_combined} report full results. LONGINUS achieves the best transition-specific F1 among LLM-based methods in 22 of 24 full-cohort settings and in 37 of 40 depth-stratified columns, and the best Macro-F1 in 14 of 16 label-frequency columns. History-based rules nonetheless lead on MIMIC-III tail labels, suggesting that many rare next-visit diagnoses are carried over from prior visits.

\begin{table*}[h!]
\centering
\scriptsize
\setlength{\tabcolsep}{5pt}
\renewcommand{\arraystretch}{1.1}

\begin{tabular}{@{}lllcccccccccccc@{}}
\toprule
\midrule
\multirow{3}{*}{\textbf{Backbone}} &
\multirow{3}{*}{\textbf{Type}} &
\multirow{3}{*}{\textbf{Method}} &
\multicolumn{6}{c}{\textbf{MIMIC-III}} &
\multicolumn{6}{c}{\textbf{MIMIC-IV}} \\
\cmidrule(lr){4-9}\cmidrule(lr){10-15}
& & &
\textbf{Agg.} & \multicolumn{2}{c}{\textbf{Dist.}} & \multicolumn{3}{c}{\textbf{Temp.}} &
\textbf{Agg.} & \multicolumn{2}{c}{\textbf{Dist.}} & \multicolumn{3}{c}{\textbf{Temp.}} \\
\cmidrule(lr){4-4}\cmidrule(lr){5-6}\cmidrule(lr){7-9}
\cmidrule(lr){10-10}\cmidrule(lr){11-12}\cmidrule(lr){13-15}
& & &
\textbf{Mic.} & \textbf{Mac.} & \textbf{Jac.} & \textbf{Pers.} & \textbf{New} & \textbf{Rec.} &
\textbf{Mic.} & \textbf{Mac.} & \textbf{Jac.} & \textbf{Pers.} & \textbf{New} & \textbf{Rec.} \\
\hline
\midrule

\multirow{2}{*}{-} & \multirow{2}{*}{Rule}
& LVCF
& .558 & .510 & .397 & .719 & .000 & .000
& .582 & .479 & .439 & .732 & .000 & .000 \\
& & OHU
& .553 & .510 & .391 & .719 & .000 & .383
& .582 & .492 & .436 & .732 & .000 & .444 \\
\midrule

\multirow{10}{*}{\makecell{Gemini-2.5\\-Flash-Lite}}
& Single-Prompting & NECHOv3
& \belowrule{.473} & \belowrule{.404} & \belowrule{.319} & \belowrule{.693} & .172 & .391
& \belowrule{.500} & \belowrule{.440} & \belowrule{.353} & \belowrule{.726} & .193 & .484 \\
\cmidrule(lr){2-15}
& \multirow{3}{*}{Multi-agent} & MedAgents
& \belowrule{.436} & \belowrule{.338} & \belowrule{.285} & \belowrule{.606} & .176 & \belowrule{.314}
& \belowrule{.484} & \belowrule{.383} & \belowrule{.336} & \belowrule{.649} & .227 & \belowrule{.400} \\
& & MDAgents
& \belowrule{.470} & \belowrule{.368} & \belowrule{.314} & \belowrule{.657} & .206 & \belowrule{.315}
& \belowrule{.509} & \belowrule{.415} & \belowrule{.352} & \belowrule{.696} & .246 & \belowrule{.418} \\
& & KAMAC
& \belowrule{.485} & \belowrule{.379} & \belowrule{.327} & \belowrule{.662} & .189 & \belowrule{.314}
& \belowrule{.527} & \belowrule{.417} & \belowrule{.372} & \belowrule{.694} & .255 & \belowrule{.433} \\
\cmidrule(lr){2-15}
& \multirow{4}{*}{Retrieval} & Summary-RAG
& \belowrule{.468} & \belowrule{.387} & \belowrule{.311} & \belowrule{.674} & \underline{.218} & .384
& \belowrule{.496} & \belowrule{.384} & \belowrule{.345} & \belowrule{.717} & .258 & .466 \\
& & RGAR
& \belowrule{.475} & \belowrule{.390} & \belowrule{.311} & \belowrule{.674} & .203 & \belowrule{.351}
& \belowrule{.513} & \belowrule{.424} & \belowrule{.347} & \belowrule{.701} & .261 & \belowrule{.444} \\
& & MASS-RAG
& \belowrule{.528} & \belowrule{\textbf{.490}} & \belowrule{.363} & \belowrule{\textbf{.718}} & .209 & .393
& \belowrule{.564} & \underline{.496} & \belowrule{.398} & .739 & \underline{.277} & .464 \\
& & EHR-RAG
& \belowrule{\textbf{.529}} & \belowrule{.446} & \belowrule{\textbf{.372}} & \belowrule{.707} & .165 & \belowrule{.322}
& \belowrule{\underline{.566}} & \belowrule{.454} & \belowrule{\underline{.416}} & \belowrule{.732} & .234 & \belowrule{.437} \\
\cmidrule(lr){2-15}
& Memory & Traj-CoA
& \belowrule{.485} & \belowrule{.399} & \belowrule{.329} & \belowrule{.699} & .210 & \underline{.402}
& \belowrule{.525} & \belowrule{.436} & \belowrule{.368} & \underline{.741} & .264 & \underline{.495} \\
\cmidrule(lr){2-15}
& Ours & \textbf{LONGINUS}
& \belowrule{\underline{.529}} & \belowrule{\underline{.467}} & \belowrule{\underline{.367}} & \belowrule{\underline{.718}} & \textbf{.226} & \textbf{.404}
& \belowrule{\textbf{.580}} & \textbf{.501} & \belowrule{\textbf{.424}} & \textbf{.756}$^{\dagger}$ & \textbf{.294}$^{\dagger}$ & \textbf{.514} \\
\midrule

\multirow{10}{*}{\makecell{Gemini-3.1\\-Flash-Lite}}
& Single-Prompting & NECHOv3
& \belowrule{.514} & \belowrule{.450} & \belowrule{.353} & \belowrule{.715} & .194 & .393
& \belowrule{.552} & \belowrule{.459} & \belowrule{.392} & \underline{.756} & .236 & .453 \\
\cmidrule(lr){2-15}
& \multirow{3}{*}{Multi-agent} & MedAgents
& \belowrule{.506} & \belowrule{.407} & \belowrule{.347} & \belowrule{.695} & .217 & \belowrule{.383}
& \belowrule{.547} & \belowrule{.444} & \belowrule{.390} & .736 & .264 & \belowrule{.416} \\
& & MDAgents
& \belowrule{.505} & \belowrule{.443} & \belowrule{.341} & \belowrule{.700} & .210 & \belowrule{.368}
& \belowrule{.546} & \belowrule{.474} & \belowrule{.387} & .736 & .263 & .478 \\
& & KAMAC
& \belowrule{.485} & \belowrule{.379} & \belowrule{.324} & \belowrule{.670} & .200 & \belowrule{.372}
& \belowrule{.525} & \belowrule{.422} & \belowrule{.365} & \belowrule{.703} & .270 & \belowrule{.435} \\
\cmidrule(lr){2-15}
& \multirow{4}{*}{Retrieval} & Summary-RAG
& \belowrule{.487} & \belowrule{.405} & \belowrule{.330} & \belowrule{.693} & .222 & \underline{.409}
& \belowrule{.518} & \belowrule{.437} & \belowrule{.356} & \belowrule{.728} & .267 & \underline{.491} \\
& & RGAR
& \belowrule{.472} & \belowrule{.391} & \belowrule{.318} & \belowrule{.674} & .196 & \belowrule{.374}
& \belowrule{.512} & \belowrule{.426} & \belowrule{.359} & \belowrule{.705} & .254 & .473 \\
& & MASS-RAG
& \belowrule{.526} & \belowrule{.454} & \belowrule{\underline{.368}} & \belowrule{.713} & .193 & .383
& \belowrule{.578} & \belowrule{.482} & \belowrule{.418} & .746 & .265 & .487 \\
& & EHR-RAG
& \belowrule{\underline{.530}} & \belowrule{\underline{.460}} & \belowrule{.367} & \belowrule{\underline{.716}} & \underline{.224} & .383
& \belowrule{\underline{.582}} & \belowrule{\underline{.484}} & \belowrule{\textbf{.420}} & .749 & \underline{.298} & .488 \\
\cmidrule(lr){2-15}
& Memory & Traj-CoA
& \belowrule{.484} & \belowrule{.402} & \belowrule{.332} & \belowrule{.691} & .184 & \belowrule{.372}
& \belowrule{.525} & \belowrule{.428} & \belowrule{.380} & \belowrule{.722} & .237 & .455 \\
\cmidrule(lr){2-15}
& Ours & \textbf{LONGINUS}
& \belowrule{\textbf{.535}} & \belowrule{\textbf{.462}} & \belowrule{\textbf{.370}} & \textbf{.728}$^{\dagger}$ & \textbf{.255}$^{\dagger}$ & \textbf{.439}
& \textbf{.583} & \belowrule{\textbf{.490}} & \belowrule{\underline{.418}} & \textbf{.773}$^{\dagger}$ & \textbf{.314} & \textbf{.529} \\
\midrule

\multirow{10}{*}{\makecell{GPT-OSS\\-120B}}
& Single-Prompting & NECHOv3
& \belowrule{\underline{.530}} & \belowrule{\underline{.451}} & \belowrule{.365} & \belowrule{.703} & .219 & .397
& \belowrule{.563} & \belowrule{.433} & \belowrule{.396} & \belowrule{.720} & .279 & .472 \\
\cmidrule(lr){2-15}
& \multirow{3}{*}{Multi-agent} & MedAgents
& \belowrule{.503} & \belowrule{.418} & \belowrule{.333} & \belowrule{.687} & \underline{.234} & \belowrule{.379}
& \belowrule{.539} & \belowrule{.439} & \belowrule{.361} & \belowrule{.715} & .274 & .456 \\
& & MDAgents
& \belowrule{.514} & \belowrule{.428} & \belowrule{.351} & \belowrule{\underline{.705}} & .232 & \belowrule{.373}
& \belowrule{.550} & \belowrule{.461} & \belowrule{.382} & .735 & .280 & .463 \\
& & KAMAC
& \belowrule{.497} & \belowrule{.383} & \belowrule{.334} & \belowrule{.674} & .207 & .399
& \belowrule{.539} & \belowrule{.429} & \belowrule{.373} & \belowrule{.713} & .272 & \belowrule{.433} \\
\cmidrule(lr){2-15}
& \multirow{4}{*}{Retrieval} & Summary-RAG
& \belowrule{.487} & \belowrule{.387} & \belowrule{.332} & \belowrule{.665} & .216 & \textbf{.421}
& \belowrule{.554} & \belowrule{.460} & \belowrule{.390} & .745 & .309 & \underline{.520} \\
& & RGAR
& \belowrule{.529} & \belowrule{.433} & \belowrule{.366} & \belowrule{.699} & .192 & \belowrule{.377}
& \belowrule{.550} & \belowrule{.443} & \belowrule{.391} & \belowrule{.731} & \underline{.310} & .492 \\
& & MASS-RAG
& \belowrule{.518} & \belowrule{.447} & \belowrule{.354} & \belowrule{.694} & .223 & .393
& \belowrule{.571} & \belowrule{\underline{.489}} & \belowrule{.411} & .740 & .289 & .450 \\
& & EHR-RAG
& \belowrule{.530} & \belowrule{.435} & \belowrule{\underline{.369}} & \belowrule{.698} & .159 & \belowrule{.367}
& \belowrule{\underline{.582}} & \belowrule{.482} & \belowrule{\textbf{.431}} & .734 & .252 & .450 \\
\cmidrule(lr){2-15}
& Memory & Traj-CoA
& \belowrule{.512} & \belowrule{.431} & \belowrule{.355} & \belowrule{.699} & .218 & \belowrule{.366}
& \belowrule{.565} & \belowrule{.458} & \belowrule{.411} & \underline{.746} & .285 & .483 \\
\cmidrule(lr){2-15}
& Ours & \textbf{LONGINUS}
& \belowrule{\textbf{.541}}$^{\dagger}$ & \belowrule{\textbf{.476}} & \belowrule{\textbf{.377}} & \textbf{.720}$^{\dagger}$ & \textbf{.257}$^{\dagger}$ & \underline{.409}
& \textbf{.587} & \textbf{.499} & \belowrule{\underline{.421}} & \textbf{.756} & \textbf{.321} & \textbf{.522} \\
\midrule

\multirow{10}{*}{\makecell{Llama-3.3\\-70B}}
& Single-Prompting & NECHOv3
& \belowrule{.510} & \belowrule{.405} & \belowrule{.354} & \belowrule{.686} & .021 & \belowrule{.297}
& \belowrule{.548} & \belowrule{.405} & \belowrule{.401} & \belowrule{.719} & .067 & \belowrule{.406} \\
\cmidrule(lr){2-15}
& \multirow{3}{*}{Multi-agent} & MedAgents
& \belowrule{.428} & \belowrule{.302} & \belowrule{.284} & \belowrule{.606} & .109 & \belowrule{.241}
& \belowrule{.468} & \belowrule{.325} & \belowrule{.330} & \belowrule{.630} & .167 & \belowrule{.378} \\
& & MDAgents
& \belowrule{.449} & \belowrule{.328} & \belowrule{.303} & \belowrule{.632} & .105 & \belowrule{.248}
& \belowrule{.487} & \belowrule{.352} & \belowrule{.348} & \belowrule{.656} & .170 & \belowrule{.344} \\
& & KAMAC
& \belowrule{.465} & \belowrule{.357} & \belowrule{.315} & \belowrule{.643} & .112 & \belowrule{.250}
& \belowrule{.502} & \belowrule{.389} & \belowrule{.362} & \belowrule{.668} & .184 & \belowrule{.358} \\
\cmidrule(lr){2-15}
& \multirow{4}{*}{Retrieval} & Summary-RAG
& \belowrule{.390} & \belowrule{.293} & \belowrule{.261} & \belowrule{.557} & .127 & \belowrule{.268}
& \belowrule{.484} & \belowrule{.378} & \belowrule{.342} & \belowrule{.676} & \underline{.226} & \belowrule{.436} \\
& & RGAR
& \belowrule{.463} & \belowrule{.358} & \belowrule{.321} & \belowrule{.642} & .056 & \belowrule{.196}
& \belowrule{.491} & \belowrule{.372} & \belowrule{.370} & \belowrule{.657} & .127 & \belowrule{.269} \\
& & MASS-RAG
& \belowrule{\underline{.518}} & \belowrule{.434} & \belowrule{\underline{.360}} & \belowrule{\underline{.694}} & .102 & \belowrule{\underline{.360}}
& \belowrule{\underline{.562}} & \belowrule{\underline{.455}} & \belowrule{.415} & \belowrule{\underline{.723}} & .195 & .450 \\
& & EHR-RAG
& \belowrule{.508} & \belowrule{\underline{.437}} & \belowrule{.357} & \belowrule{.683} & .082 & \belowrule{.160}
& \belowrule{.556} & \belowrule{.446} & \belowrule{\underline{.425}} & \belowrule{.717} & .134 & \belowrule{.350} \\
\cmidrule(lr){2-15}
& Memory & Traj-CoA
& \belowrule{.457} & \belowrule{.362} & \belowrule{.313} & \belowrule{.632} & \underline{.139} & \belowrule{.345}
& \belowrule{.502} & \belowrule{.396} & \belowrule{.364} & \belowrule{.663} & .220 & \underline{.453} \\
\cmidrule(lr){2-15}
& Ours & \textbf{LONGINUS}
& \belowrule{\textbf{.547}}$^{\dagger}$ & \belowrule{\textbf{.470}}$^{\dagger}$ & \belowrule{\textbf{.388}}$^{\dagger}$ & \textbf{.725}$^{\dagger}$ & \textbf{.174}$^{\dagger}$ & \belowrule{\textbf{.366}}
& \textbf{.597}$^{\dagger}$ & \belowrule{\textbf{.488}}$^{\dagger}$ & \textbf{.447}$^{\dagger}$ & \textbf{.759}$^{\dagger}$ & \textbf{.253}$^{\dagger}$ & \textbf{.481} \\

\bottomrule
\end{tabular}

\caption{
Full-cohort performance on MIMIC-III and MIMIC-IV for all four backbones. Formatting and $\dagger$ follow Table~\ref{tab:main_results}.
}

\label{tab:appendix_main_results}
\end{table*}

\begin{table*}[h!]
\centering
\scriptsize
\setlength{\tabcolsep}{5pt}
\renewcommand{\arraystretch}{1.1}

\begin{tabular}{@{}lllccccccccccc@{}}
\toprule
\midrule
\multirow{3}{*}{\textbf{Backbone}} &
\multirow{3}{*}{\textbf{Type}} &
\multirow{3}{*}{\textbf{Method}} &
\multicolumn{5}{c}{\textbf{Visit 1}} &
\multicolumn{6}{c}{\textbf{Visit 2+}} \\
\cmidrule(lr){4-8}\cmidrule(lr){9-14}
& & &
\textbf{Agg.} & \multicolumn{2}{c}{\textbf{Dist.}} & \multicolumn{2}{c}{\textbf{Temp.}} &
\textbf{Agg.} & \multicolumn{2}{c}{\textbf{Dist.}} & \multicolumn{3}{c}{\textbf{Temp.}} \\
\cmidrule(lr){4-4}\cmidrule(lr){5-6}\cmidrule(lr){7-8}
\cmidrule(lr){9-9}\cmidrule(lr){10-11}\cmidrule(lr){12-14}
& & &
\textbf{Mic.} & \textbf{Mac.} & \textbf{Jac.} & \textbf{Pers.} & \textbf{New} &
\textbf{Mic.} & \textbf{Mac.} & \textbf{Jac.} & \textbf{Pers.} & \textbf{New} & \textbf{Rec.} \\
\hline
\midrule

\multirow{2}{*}{-} & \multirow{2}{*}{Rule}
& LVCF
& .548 & .489 & .389 & .712 & .000
& .591 & .467 & .426 & .741 & .000 & .000 \\
& & OHU
& .548 & .489 & .389 & .712 & .000
& .566 & .467 & .397 & .741 & .000 & .383 \\
\midrule

\multirow{10}{*}{\makecell{Gemini-2.5\\-Flash-Lite}}
& Single-Prompting & NECHOv3
& \belowrule{.461} & \belowrule{.391} & \belowrule{.309} & \belowrule{.684} & .179
& \belowrule{.508} & \belowrule{.384} & \belowrule{.355} & \belowrule{.717} & .140 & .391 \\
\cmidrule(lr){2-14}
& \multirow{3}{*}{Multi-agent} & MedAgents
& \belowrule{.446} & \belowrule{.345} & \belowrule{.294} & \belowrule{.623} & .185
& \belowrule{.404} & \belowrule{.269} & \belowrule{.251} & \belowrule{.549} & .133 & \belowrule{.314} \\
& & MDAgents
& \belowrule{.475} & \belowrule{.378} & \belowrule{.319} & \belowrule{.664} & .218
& \belowrule{.454} & \belowrule{.288} & \belowrule{.297} & \belowrule{.635} & .144 & \belowrule{.315} \\
& & KAMAC
& \belowrule{.489} & \belowrule{.388} & \belowrule{.330} & \belowrule{.668} & .194
& \belowrule{.475} & \belowrule{.284} & \belowrule{.316} & \belowrule{.645} & .161 & \belowrule{.314} \\
\cmidrule(lr){2-14}
& \multirow{4}{*}{Retrieval} & Summary-RAG
& \belowrule{.475} & \belowrule{.395} & \belowrule{.319} & \belowrule{.685} & \underline{.231}
& \belowrule{.444} & \belowrule{.321} & \belowrule{.282} & \belowrule{.638} & .157 & .384 \\
& & RGAR
& \belowrule{.478} & \belowrule{.393} & \belowrule{.315} & \belowrule{.675} & .214
& \belowrule{.462} & \belowrule{.344} & \belowrule{.297} & \belowrule{.669} & .153 & \belowrule{.351} \\
& & MASS-RAG
& \belowrule{.522} & \belowrule{\textbf{.466}} & \belowrule{.357} & \belowrule{\underline{.710}} & .219
& \belowrule{\textbf{.549}} & \belowrule{\textbf{.455}} & \belowrule{.384} & \textbf{.745} & .147 & .393 \\
& & EHR-RAG
& \belowrule{\textbf{.523}} & \belowrule{.440} & \belowrule{\textbf{.366}} & \belowrule{.702} & .175
& \belowrule{\underline{.548}} & \belowrule{.405} & \belowrule{\textbf{.394}} & \belowrule{.725} & .116 & \belowrule{.322} \\
\cmidrule(lr){2-14}
& Memory & Traj-CoA
& \belowrule{.483} & \belowrule{.392} & \belowrule{.329} & \belowrule{.699} & .218
& \belowrule{.492} & \belowrule{.381} & \belowrule{.332} & \belowrule{.701} & \textbf{.177} & \underline{.402} \\
\cmidrule(lr){2-14}
& Ours & \textbf{LONGINUS}
& \belowrule{\underline{.523}} & \belowrule{\underline{.447}} & \belowrule{\underline{.362}} & \belowrule{\textbf{.710}} & \textbf{.236}
& \belowrule{.546} & \belowrule{\underline{.445}} & \belowrule{\underline{.384}} & \belowrule{\underline{.741}} & \underline{.175} & \textbf{.404} \\
\midrule

\multirow{10}{*}{\makecell{Gemini-3.1\\-Flash-Lite}}
& Single-Prompting & NECHOv3
& \belowrule{.506} & \belowrule{.433} & \belowrule{.347} & \belowrule{\underline{.711}} & .203
& \belowrule{.539} & \belowrule{.426} & \belowrule{.376} & \belowrule{.729} & .146 & .393 \\
\cmidrule(lr){2-14}
& \multirow{3}{*}{Multi-agent} & MedAgents
& \belowrule{.504} & \belowrule{.396} & \belowrule{.345} & \belowrule{.694} & .229
& \belowrule{.509} & \belowrule{.350} & \belowrule{.351} & \belowrule{.696} & .161 & \belowrule{.383} \\
& & MDAgents
& \belowrule{.505} & \belowrule{\underline{.441}} & \belowrule{.342} & \belowrule{.699} & .223
& \belowrule{.505} & \belowrule{.382} & \belowrule{.338} & \belowrule{.701} & .145 & \belowrule{.368} \\
& & KAMAC
& \belowrule{.487} & \belowrule{.380} & \belowrule{.327} & \belowrule{.673} & .214
& \belowrule{.479} & \belowrule{.327} & \belowrule{.316} & \belowrule{.662} & .135 & \belowrule{.372} \\
\cmidrule(lr){2-14}
& \multirow{4}{*}{Retrieval} & Summary-RAG
& \belowrule{.480} & \belowrule{.397} & \belowrule{.325} & \belowrule{.688} & .229
& \belowrule{.509} & \belowrule{.387} & \belowrule{.348} & \belowrule{.707} & \underline{.189} & \underline{.409} \\
& & RGAR
& \belowrule{.467} & \belowrule{.387} & \belowrule{.314} & \belowrule{.673} & .204
& \belowrule{.488} & \belowrule{.356} & \belowrule{.331} & \belowrule{.676} & .152 & \belowrule{.374} \\
& & MASS-RAG
& \belowrule{.521} & \belowrule{.436} & \belowrule{\underline{.365}} & \belowrule{.710} & .206
& \belowrule{.541} & \belowrule{.421} & \belowrule{.379} & \belowrule{.722} & .126 & .383 \\
& & EHR-RAG
& \belowrule{\underline{.524}} & \belowrule{\textbf{.442}} & \belowrule{.363} & \belowrule{.709} & \underline{.237}
& \belowrule{\underline{.548}} & \belowrule{\underline{.435}} & \belowrule{\underline{.381}} & \belowrule{\underline{.737}} & .155 & .383 \\
\cmidrule(lr){2-14}
& Memory & Traj-CoA
& \belowrule{.483} & \belowrule{.402} & \belowrule{.332} & \belowrule{.691} & .196
& \belowrule{.487} & \belowrule{.360} & \belowrule{.330} & \belowrule{.689} & .126 & \belowrule{.372} \\
\cmidrule(lr){2-14}
& Ours & \textbf{LONGINUS}
& \belowrule{\textbf{.529}} & \belowrule{.438} & \belowrule{\textbf{.365}} & \textbf{.723}$^{\dagger}$ & \textbf{.266}$^{\dagger}$
& \belowrule{\textbf{.553}} & \belowrule{\textbf{.448}} & \belowrule{\textbf{.387}} & \belowrule{\textbf{.741}} & \textbf{.199} & \textbf{.439} \\
\midrule

\multirow{10}{*}{\makecell{GPT-OSS\\-120B}}
& Single-Prompting & NECHOv3
& \belowrule{.521} & \belowrule{.439} & \belowrule{.359} & \belowrule{.698} & .225
& \belowrule{\textbf{.556}} & \belowrule{\underline{.424}} & \belowrule{\underline{.387}} & \belowrule{\underline{.718}} & .187 & .397 \\
\cmidrule(lr){2-14}
& \multirow{3}{*}{Multi-agent} & MedAgents
& \belowrule{.512} & \belowrule{.426} & \belowrule{.344} & \belowrule{.696} & \underline{.250}
& \belowrule{.474} & \belowrule{.348} & \belowrule{.295} & \belowrule{.657} & .162 & \belowrule{.379} \\
& & MDAgents
& \belowrule{.520} & \belowrule{.436} & \belowrule{.357} & \belowrule{.707} & .247
& \belowrule{.496} & \belowrule{.363} & \belowrule{.327} & \belowrule{.698} & .165 & \belowrule{.373} \\
& & KAMAC
& \belowrule{.497} & \belowrule{.382} & \belowrule{.335} & \belowrule{.676} & .218
& \belowrule{.497} & \belowrule{.330} & \belowrule{.329} & \belowrule{.669} & .155 & .399 \\
\cmidrule(lr){2-14}
& \multirow{4}{*}{Retrieval} & Summary-RAG
& \belowrule{.482} & \belowrule{.377} & \belowrule{.329} & \belowrule{.664} & .222
& \belowrule{.505} & \belowrule{.361} & \belowrule{.344} & \belowrule{.669} & \underline{.189} & \textbf{.421} \\
& & RGAR
& \belowrule{\underline{.527}} & \belowrule{.429} & \belowrule{.364} & \belowrule{.699} & .203
& \belowrule{.535} & \belowrule{.370} & \belowrule{.372} & \belowrule{.700} & .128 & \belowrule{.377} \\
& & MASS-RAG
& \belowrule{.519} & \belowrule{\underline{.442}} & \belowrule{.358} & \belowrule{.696} & .234
& \belowrule{.518} & \belowrule{.405} & \belowrule{.342} & \belowrule{.688} & .160 & .393 \\
& & EHR-RAG
& \belowrule{.522} & \belowrule{.426} & \belowrule{.363} & \belowrule{.694} & .167
& \belowrule{\underline{.555}} & \belowrule{.390} & \belowrule{\textbf{.391}} & \belowrule{.710} & .111 & \belowrule{.367} \\
\cmidrule(lr){2-14}
& Memory & Traj-CoA
& \belowrule{.520} & \belowrule{.417} & \belowrule{\underline{.365}} & \belowrule{\underline{.710}} & .231
& \belowrule{.485} & \belowrule{.377} & \belowrule{.318} & \belowrule{.664} & .152 & \belowrule{.366} \\
\cmidrule(lr){2-14}
& Ours & \textbf{LONGINUS}
& \belowrule{\textbf{.537}} & \belowrule{\textbf{.460}} & \belowrule{\textbf{.375}} & \textbf{.714} & \textbf{.268}
& \belowrule{.553} & \belowrule{\textbf{.433}} & \belowrule{.386} & \belowrule{\textbf{.738}}$^{\dagger}$ & \textbf{.199} & \underline{.409} \\
\midrule

\multirow{10}{*}{\makecell{Llama-3.3\\-70B}}
& Single-Prompting & NECHOv3
& \belowrule{.494} & \belowrule{.379} & \belowrule{.340} & \belowrule{.675} & .022
& \belowrule{\underline{.560}} & \belowrule{\underline{.400}} & \belowrule{\underline{.402}} & \belowrule{\underline{.721}} & .012 & \belowrule{.297} \\
\cmidrule(lr){2-14}
& \multirow{3}{*}{Multi-agent} & MedAgents
& \belowrule{.430} & \belowrule{.300} & \belowrule{.287} & \belowrule{.614} & .118
& \belowrule{.420} & \belowrule{.241} & \belowrule{.273} & \belowrule{.583} & .058 & \belowrule{.241} \\
& & MDAgents
& \belowrule{.447} & \belowrule{.324} & \belowrule{.302} & \belowrule{.635} & .114
& \belowrule{.456} & \belowrule{.281} & \belowrule{.304} & \belowrule{.622} & .052 & \belowrule{.248} \\
& & KAMAC
& \belowrule{.464} & \belowrule{.357} & \belowrule{.316} & \belowrule{.646} & .122
& \belowrule{.467} & \belowrule{.292} & \belowrule{.311} & \belowrule{.635} & .054 & \belowrule{.250} \\
\cmidrule(lr){2-14}
& \multirow{4}{*}{Retrieval} & Summary-RAG
& \belowrule{.389} & \belowrule{.281} & \belowrule{.262} & \belowrule{.561} & .134
& \belowrule{.391} & \belowrule{.264} & \belowrule{.257} & \belowrule{.543} & .083 & \belowrule{.268} \\
& & RGAR
& \belowrule{.473} & \belowrule{.365} & \belowrule{.331} & \belowrule{.655} & .065
& \belowrule{.428} & \belowrule{.268} & \belowrule{.283} & \belowrule{.592} & .004 & \belowrule{.196} \\
& & MASS-RAG
& \belowrule{\underline{.510}} & \belowrule{\underline{.425}} & \belowrule{\underline{.355}} & \belowrule{\underline{.689}} & .113
& \belowrule{.541} & \belowrule{.388} & \belowrule{.378} & \belowrule{.710} & .036 & \belowrule{\underline{.360}} \\
& & EHR-RAG
& \belowrule{.501} & \belowrule{.417} & \belowrule{.353} & \belowrule{.680} & .088
& \belowrule{.530} & \belowrule{.394} & \belowrule{.371} & \belowrule{.694} & .050 & \belowrule{.160} \\
\cmidrule(lr){2-14}
& Memory & Traj-CoA
& \belowrule{.450} & \belowrule{.347} & \belowrule{.310} & \belowrule{.633} & \underline{.146}
& \belowrule{.477} & \belowrule{.327} & \belowrule{.327} & \belowrule{.631} & \underline{.100} & \belowrule{.345} \\
\cmidrule(lr){2-14}
& Ours & \textbf{LONGINUS}
& \belowrule{\textbf{.538}}$^{\dagger}$ & \belowrule{\textbf{.450}}$^{\dagger}$ & \belowrule{\textbf{.382}}$^{\dagger}$ & \textbf{.718}$^{\dagger}$ & \textbf{.186}$^{\dagger}$
& \belowrule{\textbf{.576}} & \belowrule{\textbf{.416}} & \belowrule{\textbf{.410}} & \textbf{.746} & \textbf{.107} & \belowrule{\textbf{.366}} \\

\bottomrule
\end{tabular}

\caption{
Performance on MIMIC-III by observed history depth. Visit~1 and Visit~2+ contain 786 and 214 patients, respectively; Recurrent is undefined for Visit~1. Formatting and $\dagger$ follow Table~\ref{tab:main_results}.
}
\label{tab:full_history_mimic_iii}
\end{table*}

\begin{table*}[h!]
\centering
\scriptsize
\setlength{\tabcolsep}{5pt}
\renewcommand{\arraystretch}{1.1}

\begin{tabular}{@{}lllccccccccccc@{}}
\toprule
\midrule
\multirow{3}{*}{\textbf{Backbone}} &
\multirow{3}{*}{\textbf{Type}} &
\multirow{3}{*}{\textbf{Method}} &
\multicolumn{5}{c}{\textbf{Visit 1}} &
\multicolumn{6}{c}{\textbf{Visit 2+}} \\
\cmidrule(lr){4-8}\cmidrule(lr){9-14}
& & &
\textbf{Agg.} & \multicolumn{2}{c}{\textbf{Dist.}} & \multicolumn{2}{c}{\textbf{Temp.}} &
\textbf{Agg.} & \multicolumn{2}{c}{\textbf{Dist.}} & \multicolumn{3}{c}{\textbf{Temp.}} \\
\cmidrule(lr){4-4}\cmidrule(lr){5-6}\cmidrule(lr){7-8}
\cmidrule(lr){9-9}\cmidrule(lr){10-11}\cmidrule(lr){12-14}
& & &
\textbf{Mic.} & \textbf{Mac.} & \textbf{Jac.} & \textbf{Pers.} & \textbf{New} &
\textbf{Mic.} & \textbf{Mac.} & \textbf{Jac.} & \textbf{Pers.} & \textbf{New} & \textbf{Rec.} \\
\hline
\midrule

\multirow{2}{*}{-} & \multirow{2}{*}{Rule}
& LVCF
& .582 & .470 & .440 & .732 & .000
& .586 & .402 & .437 & .734 & .000 & .000 \\
& & OHU
& .582 & .470 & .440 & .732 & .000
& .582 & .455 & .426 & .734 & .000 & .444 \\
\midrule

\multirow{10}{*}{\makecell{Gemini-2.5\\-Flash-Lite}}
& Single-Prompting & NECHOv3
& \belowrule{.493} & \belowrule{.409} & \belowrule{.344} & \belowrule{.722} & .203
& \belowrule{.521} & \belowrule{.415} & \belowrule{.377} & .741 & .148 & .484 \\
\cmidrule(lr){2-14}
& \multirow{3}{*}{Multi-agent} & MedAgents
& \belowrule{.489} & \belowrule{.374} & \belowrule{.340} & \belowrule{.656} & .241
& \belowrule{.465} & \belowrule{.321} & \belowrule{.323} & \belowrule{.627} & .156 & \belowrule{.400} \\
& & MDAgents
& \belowrule{.513} & \belowrule{.408} & \belowrule{.354} & \belowrule{.697} & .263
& \belowrule{.493} & \belowrule{.360} & \belowrule{.347} & \belowrule{.690} & .162 & \belowrule{.418} \\
& & KAMAC
& \belowrule{.527} & \belowrule{.398} & \belowrule{.373} & \belowrule{.693} & .264
& \belowrule{.526} & \belowrule{.383} & \belowrule{.368} & \belowrule{.696} & .212 & \belowrule{.433} \\
\cmidrule(lr){2-14}
& \multirow{4}{*}{Retrieval} & Summary-RAG
& \belowrule{.501} & \belowrule{.373} & \belowrule{.357} & \belowrule{.718} & .275
& \belowrule{.480} & \belowrule{.359} & \belowrule{.311} & \belowrule{.714} & .184 & .466 \\
& & RGAR
& \belowrule{.517} & \belowrule{.423} & \belowrule{.357} & \belowrule{.701} & .274
& \belowrule{.499} & \belowrule{.367} & \belowrule{.318} & \belowrule{.702} & .203 & \belowrule{.444} \\
& & MASS-RAG
& \belowrule{\underline{.566}} & \textbf{.481} & \belowrule{.402} & .739 & \underline{.289}
& \belowrule{.558} & \belowrule{\underline{.451}} & \belowrule{.387} & .740 & \underline{.217} & .464 \\
& & EHR-RAG
& \belowrule{.564} & \belowrule{.438} & \belowrule{\underline{.416}} & \belowrule{.731} & .247
& \belowrule{\underline{.574}} & \belowrule{.421} & \belowrule{\underline{.414}} & .737 & .157 & \belowrule{.437} \\
\cmidrule(lr){2-14}
& Memory & Traj-CoA
& \belowrule{.531} & \belowrule{.431} & \belowrule{.375} & \underline{.741} & .283
& \belowrule{.507} & \belowrule{.374} & \belowrule{.346} & \underline{.743} & .186 & \underline{.495} \\
\cmidrule(lr){2-14}
& Ours & \textbf{LONGINUS}
& \belowrule{\textbf{.578}}$^{\dagger}$ & \underline{.477} & \belowrule{\textbf{.424}} & \textbf{.756}$^{\dagger}$ & \textbf{.304}
& \textbf{.589} & \textbf{.461} & \belowrule{\textbf{.424}} & \textbf{.757} & \textbf{.245} & \textbf{.514} \\
\midrule

\multirow{10}{*}{\makecell{Gemini-3.1\\-Flash-Lite}}
& Single-Prompting & NECHOv3
& \belowrule{.552} & \belowrule{.435} & \belowrule{.391} & \underline{.752} & .254
& \belowrule{.552} & \belowrule{.411} & \belowrule{.396} & \underline{.768} & .153 & .453 \\
\cmidrule(lr){2-14}
& \multirow{3}{*}{Multi-agent} & MedAgents
& \belowrule{.552} & \belowrule{.426} & \belowrule{.397} & .736 & .279
& \belowrule{.532} & \belowrule{.388} & \belowrule{.372} & \belowrule{.734} & .195 & \belowrule{.416} \\
& & MDAgents
& \belowrule{.549} & \belowrule{.457} & \belowrule{.392} & .733 & .283
& \belowrule{.537} & \belowrule{.404} & \belowrule{.375} & .746 & .174 & .478 \\
& & KAMAC
& \belowrule{.528} & \belowrule{.406} & \belowrule{.371} & \belowrule{.700} & .286
& \belowrule{.514} & \belowrule{.385} & \belowrule{.349} & \belowrule{.710} & .197 & \belowrule{.435} \\
\cmidrule(lr){2-14}
& \multirow{4}{*}{Retrieval} & Summary-RAG
& \belowrule{.522} & \belowrule{.425} & \belowrule{.362} & \belowrule{.723} & .285
& \belowrule{.508} & \belowrule{.393} & \belowrule{.338} & .744 & .199 & \underline{.491} \\
& & RGAR
& \belowrule{.511} & \belowrule{.413} & \belowrule{.360} & \belowrule{.698} & .271
& \belowrule{.515} & \belowrule{.371} & \belowrule{.356} & \belowrule{.729} & .181 & .473 \\
& & MASS-RAG
& \belowrule{.576} & \belowrule{.458} & \belowrule{.418} & .746 & .274
& \belowrule{\textbf{.583}} & \belowrule{\textbf{.449}} & \belowrule{\textbf{.418}} & .746 & \underline{.223} & .487 \\
& & EHR-RAG
& \underline{.584} & \textbf{.483} & \belowrule{\textbf{.424}} & .748 & \underline{.315}
& \belowrule{.575} & \belowrule{.426} & \belowrule{.405} & .753 & .213 & .488 \\
\cmidrule(lr){2-14}
& Memory & Traj-CoA
& \belowrule{.526} & \belowrule{.411} & \belowrule{.384} & \belowrule{.719} & .250
& \belowrule{.525} & \belowrule{.386} & \belowrule{.369} & .735 & .182 & .455 \\
\cmidrule(lr){2-14}
& Ours & \textbf{LONGINUS}
& \textbf{.584} & \underline{.477} & \belowrule{\underline{.421}} & \textbf{.773}$^{\dagger}$ & \textbf{.330}
& \belowrule{\underline{.580}} & \belowrule{\underline{.437}} & \belowrule{\underline{.408}} & \textbf{.774} & \textbf{.234} & \textbf{.529} \\
\midrule

\multirow{10}{*}{\makecell{GPT-OSS\\-120B}}
& Single-Prompting & NECHOv3
& \belowrule{.564} & \belowrule{.429} & \belowrule{.399} & \belowrule{.722} & .290
& \belowrule{.561} & \belowrule{.395} & \belowrule{.386} & \belowrule{.713} & .223 & .472 \\
\cmidrule(lr){2-14}
& \multirow{3}{*}{Multi-agent} & MedAgents
& \belowrule{.551} & \belowrule{.440} & \belowrule{.383} & \belowrule{.727} & .290
& \belowrule{.498} & \belowrule{.357} & \belowrule{.298} & \belowrule{.673} & .196 & .456 \\
& & MDAgents
& \belowrule{.557} & \belowrule{.451} & \belowrule{.395} & .736 & .298
& \belowrule{.527} & \belowrule{.398} & \belowrule{.345} & \belowrule{.731} & .197 & .463 \\
& & KAMAC
& \belowrule{.548} & \belowrule{.419} & \belowrule{.390} & \belowrule{.722} & .284
& \belowrule{.509} & \belowrule{.385} & \belowrule{.323} & \belowrule{.681} & .215 & \belowrule{.433} \\
\cmidrule(lr){2-14}
& \multirow{4}{*}{Retrieval} & Summary-RAG
& \belowrule{.560} & \belowrule{.452} & \belowrule{.398} & \underline{.749} & \underline{.325}
& \belowrule{.537} & \belowrule{.386} & \belowrule{.368} & \belowrule{.733} & .237 & \underline{.520} \\
& & RGAR
& \belowrule{.552} & \belowrule{.434} & \belowrule{.396} & \belowrule{.731} & .323
& \belowrule{.544} & \belowrule{.380} & \belowrule{.377} & \belowrule{.734} & .255 & .492 \\
& & MASS-RAG
& \belowrule{.570} & \belowrule{.462} & \belowrule{.414} & .738 & .296
& \belowrule{.572} & \belowrule{\underline{.449}} & \belowrule{.400} & \underline{.748} & \underline{.256} & .450 \\
& & EHR-RAG
& \belowrule{\underline{.579}} & \underline{.472} & \belowrule{\textbf{.433}} & \belowrule{.732} & .265
& \textbf{.589} & \belowrule{.427} & \belowrule{\textbf{.426}} & .743 & .177 & .450 \\
\cmidrule(lr){2-14}
& Memory & Traj-CoA
& \belowrule{.571} & \belowrule{.449} & \belowrule{\underline{.422}} & .748 & .303
& \belowrule{.545} & \belowrule{.385} & \belowrule{.380} & .740 & .206 & .483 \\
\cmidrule(lr){2-14}
& Ours & \textbf{LONGINUS}
& \textbf{.587} & \textbf{.482} & \belowrule{.422} & \textbf{.758} & \textbf{.332}
& \belowrule{\underline{.585}} & \belowrule{\textbf{.454}} & \belowrule{\underline{.419}} & \textbf{.749} & \textbf{.264} & \textbf{.522} \\
\midrule

\multirow{10}{*}{\makecell{Llama-3.3\\-70B}}
& Single-Prompting & NECHOv3
& \belowrule{.533} & \belowrule{.380} & \belowrule{.383} & \belowrule{.710} & .064
& \underline{.600} & \belowrule{.379} & \underline{.453} & \underline{.747} & .082 & \belowrule{.406} \\
\cmidrule(lr){2-14}
& \multirow{3}{*}{Multi-agent} & MedAgents
& \belowrule{.466} & \belowrule{.303} & \belowrule{.329} & \belowrule{.630} & .175
& \belowrule{.476} & \belowrule{.306} & \belowrule{.332} & \belowrule{.631} & .118 & \belowrule{.378} \\
& & MDAgents
& \belowrule{.483} & \belowrule{.334} & \belowrule{.343} & \belowrule{.654} & .178
& \belowrule{.501} & \belowrule{.320} & \belowrule{.361} & \belowrule{.663} & .123 & \belowrule{.344} \\
& & KAMAC
& \belowrule{.500} & \belowrule{.367} & \belowrule{.358} & \belowrule{.666} & .195
& \belowrule{.510} & \belowrule{.348} & \belowrule{.372} & \belowrule{.678} & .122 & \belowrule{.358} \\
\cmidrule(lr){2-14}
& \multirow{4}{*}{Retrieval} & Summary-RAG
& \belowrule{.490} & \belowrule{.370} & \belowrule{.345} & \belowrule{.674} & \underline{.247}
& \belowrule{.465} & \belowrule{.322} & \belowrule{.331} & \belowrule{.685} & .144 & \belowrule{.436} \\
& & RGAR
& \belowrule{.501} & \belowrule{.367} & \belowrule{.381} & \belowrule{.669} & .140
& \belowrule{.451} & \belowrule{.289} & \belowrule{.336} & \belowrule{.608} & .034 & \belowrule{.269} \\
& & MASS-RAG
& \belowrule{\underline{.558}} & \belowrule{\underline{.433}} & \belowrule{.416} & \belowrule{\underline{.724}} & .202
& \belowrule{.572} & \belowrule{\underline{.422}} & \belowrule{.411} & \belowrule{.718} & .151 & .450 \\
& & EHR-RAG
& \belowrule{.548} & \belowrule{.424} & \belowrule{\underline{.419}} & \belowrule{.713} & .138
& \belowrule{.584} & \belowrule{.397} & .443 & \belowrule{.731} & .107 & \belowrule{.350} \\
\cmidrule(lr){2-14}
& Memory & Traj-CoA
& \belowrule{.494} & \belowrule{.374} & \belowrule{.359} & \belowrule{.656} & .227
& \belowrule{.527} & \belowrule{.356} & \belowrule{.379} & \belowrule{.686} & \underline{.184} & \underline{.453} \\
\cmidrule(lr){2-14}
& Ours & \textbf{LONGINUS}
& \textbf{.592}$^{\dagger}$ & \belowrule{\textbf{.463}}$^{\dagger}$ & \textbf{.444}$^{\dagger}$ & \textbf{.757}$^{\dagger}$ & \textbf{.262}
& \textbf{.613} & \belowrule{\textbf{.437}} & \textbf{.456} & \textbf{.764} & \textbf{.197} & \textbf{.481} \\

\bottomrule
\end{tabular}

\caption{
Performance on MIMIC-IV by observed history depth. Visit~1 and Visit~2+ contain 744 and 256 patients, respectively; Recurrent is undefined for Visit~1. Formatting and $\dagger$ follow Table~\ref{tab:main_results}.}
\label{tab:full_history_mimic_iv}
\end{table*}
\begin{table*}[t]
\centering
\scriptsize
\setlength{\tabcolsep}{5pt}
\renewcommand{\arraystretch}{1.1}

\begin{tabular}{@{}lllcccccccc@{}}
\toprule
\multirow{2}{*}{\textbf{Backbone}}
& \multirow{2}{*}{\textbf{Type}}
& \multirow{2}{*}{\textbf{Method}}
& \multicolumn{4}{c}{\textbf{MIMIC-III}}
& \multicolumn{4}{c}{\textbf{MIMIC-IV}} \\
\cmidrule(lr){4-7}
\cmidrule(lr){8-11}
& & &
\textbf{Head} & \textbf{Mid} & \textbf{Tail} & \textbf{All}
& \textbf{Head} & \textbf{Mid} & \textbf{Tail} & \textbf{All} \\
\midrule

\multirow{2}{*}{-}
& \multirow{2}{*}{Rule}
& LVCF
& .562 & .433 & .536 & .510
& .599 & .472 & .434 & .479 \\
& & OHU
& .566 & .428 & .537 & .510
& .606 & .476 & .454 & .492 \\
\midrule

\multirow{10}{*}{\makecell{Gemini-3.1\\-Flash-Lite}}
& Single-Prompting & NECHOv3
& \belowrule{.511} & \belowrule{.383}
& \belowrule{.465} & \belowrule{.450}
& \belowrule{.574} & \belowrule{.430}
& \belowrule{.428} & \belowrule{.459} \\
\cmidrule(lr){2-11}

& \multirow{3}{*}{Multi-agent} & MedAgents
& \belowrule{.503} & \belowrule{.359}
& \belowrule{.397} & \belowrule{.407}
& \belowrule{.570} & \belowrule{.418}
& \belowrule{.408} & \belowrule{.444} \\
& & MDAgents
& \belowrule{.514} & \belowrule{.379}
& \belowrule{.453} & \belowrule{.443}
& \belowrule{.574} & \belowrule{.438}
& \belowrule{\textbf{.454}} & \belowrule{.474} \\
& & KAMAC
& \belowrule{.508} & \belowrule{.346}
& \belowrule{.346} & \belowrule{.379}
& \belowrule{.560} & \belowrule{.415}
& \belowrule{.368} & \belowrule{.422} \\
\cmidrule(lr){2-11}

& \multirow{4}{*}{Retrieval} & Summary-RAG
& \belowrule{.470} & \belowrule{.368}
& \belowrule{.401} & \belowrule{.405}
& \belowrule{.539} & \belowrule{.421}
& \belowrule{.403} & \belowrule{.437} \\
& & RGAR
& \belowrule{.431} & \belowrule{.356}
& \belowrule{.395} & \belowrule{.391}
& \belowrule{.513} & \belowrule{.401}
& \belowrule{.405} & \belowrule{.426} \\
& & MASS-RAG
& \belowrule{.508} & \belowrule{.388}
& \belowrule{\textbf{.472}} & \belowrule{.454}
& \belowrule{.587} & \belowrule{.457}
& \belowrule{\underline{.454}} & \belowrule{.482} \\
& & EHR-RAG
& \belowrule{\underline{.545}} & \belowrule{\underline{.392}}
& \belowrule{\underline{.466}} & \belowrule{\underline{.460}}
& \underline{.617} & \belowrule{\underline{.465}}
& \belowrule{.440} & \belowrule{\underline{.484}} \\
\cmidrule(lr){2-11}

& Memory & Traj-CoA
& \belowrule{.431} & \belowrule{.364}
& \belowrule{.413} & \belowrule{.402}
& \belowrule{.515} & \belowrule{.410}
& \belowrule{.402} & \belowrule{.428} \\
\cmidrule(lr){2-11}

& Ours & \textbf{LONGINUS}
& \belowrule{\textbf{.557}} & \belowrule{\textbf{.396}}
& \belowrule{.463} & \belowrule{\textbf{.462}}
& \textbf{.624} & \textbf{.486}
& \belowrule{.436} & \belowrule{\textbf{.490}} \\
\midrule

\multirow{10}{*}{\makecell{Llama-3.3\\-70B}}
& Single-Prompting & NECHOv3
& \belowrule{\underline{.519}} & \belowrule{.341}
& \belowrule{.397} & \belowrule{.405}
& \belowrule{.547} & \belowrule{.381}
& \belowrule{.361} & \belowrule{.405} \\
\cmidrule(lr){2-11}

& \multirow{3}{*}{Multi-agent} & MedAgents
& \belowrule{.411} & \belowrule{.257}
& \belowrule{.284} & \belowrule{.302}
& \belowrule{.456} & \belowrule{.295}
& \belowrule{.289} & \belowrule{.325} \\
& & MDAgents
& \belowrule{.443} & \belowrule{.287}
& \belowrule{.306} & \belowrule{.328}
& \belowrule{.476} & \belowrule{.338}
& \belowrule{.308} & \belowrule{.352} \\
& & KAMAC
& \belowrule{.454} & \belowrule{.310}
& \belowrule{.346} & \belowrule{.357}
& \belowrule{.491} & \belowrule{.355}
& \belowrule{.367} & \belowrule{.389} \\
\cmidrule(lr){2-11}

& \multirow{4}{*}{Retrieval} & Summary-RAG
& \belowrule{.361} & \belowrule{.233}
& \belowrule{.301} & \belowrule{.293}
& \belowrule{.446} & \belowrule{.348}
& \belowrule{.368} & \belowrule{.378} \\
& & RGAR
& \belowrule{.454} & \belowrule{.303}
& \belowrule{.351} & \belowrule{.358}
& \belowrule{.469} & \belowrule{.323}
& \belowrule{.362} & \belowrule{.372} \\
& & MASS-RAG
& \belowrule{.510} & \belowrule{\underline{.379}}
& \belowrule{.435} & \belowrule{.434}
& \belowrule{.548} & \belowrule{\underline{.428}}
& \belowrule{\underline{.433}} & \belowrule{\underline{.455}} \\
& & EHR-RAG
& \belowrule{.495} & \belowrule{.366}
& \belowrule{\underline{.458}} & \belowrule{\underline{.437}}
& \belowrule{\underline{.555}} & \belowrule{.414}
& \belowrule{.420} & \belowrule{.446} \\
\cmidrule(lr){2-11}

& Memory & Traj-CoA
& \belowrule{.433} & \belowrule{.291}
& \belowrule{.375} & \belowrule{.362}
& \belowrule{.459} & \belowrule{.374}
& \belowrule{.384} & \belowrule{.396} \\
\cmidrule(lr){2-11}

& Ours & \textbf{LONGINUS}
& \belowrule{\textbf{.542}} & \belowrule{\textbf{.404}}
& \belowrule{\textbf{.480}} & \belowrule{\textbf{.470}}
& \belowrule{\textbf{.597}} & \belowrule{\textbf{.463}}
& \textbf{.458} & \belowrule{\textbf{.488}} \\

\bottomrule
\end{tabular}

\caption{
Macro-F1 by label-frequency stratum on MIMIC-III and MIMIC-IV (1,000 patients each; up to five observed visits). Labels are ranked by next-visit target prevalence in each dataset: Head (top 20\%; 23 labels), Mid (next 30\%; 34 labels), and Tail (remaining 50\%; 55 labels); All covers the full 112-label space. Per-stratum Macro-F1 averages per-label F1 within each stratum (Appendix~\ref{app:eval_metrics}). Formatting follows Table~\ref{tab:main_results}; no significance tests are reported for frequency strata.
}
\label{tab:freq_macro_combined}
\end{table*}

\newpage
\clearpage
\twocolumn

\section{Additional Precision--Recall Analysis}
\label{app:precision_recall}

\subsection{Overall Stage-wise Precision and Recall}

Table~\ref{tab:stagewise_pr} reports stage-wise micro-precision and recall. The raw candidate union contains predictions from both views, while the final output retains consensus and accepted disagreements. Adjudication increases precision from 43.2\% to 49.2\% with Gemini-3.1-Flash-Lite and from 52.4\% to 59.7\% with Llama-3.3-70B, while reducing recall from 76.4\% to 71.6\% and from 64.1\% to 59.8\%, respectively. This reflects the precision--recall trade-off of filtering disagreement candidates.

\subsection{Transition-Specific Precision and Recall}

Table~\ref{tab:transition_pr} reports precision and recall by temporal transition using the history-state partition from the main evaluation. Adjudication improves precision at the cost of recall across all transitions, with the largest recall losses for Recurrent diagnoses (9.6 and 11.8 points). New recall remains low even before adjudication (46.2\% and 26.7\%), and New F1 improves with Gemini-3.1-Flash-Lite but slightly declines with Llama-3.3-70B.

\begin{table}[h]
\centering
\scriptsize
\setlength{\tabcolsep}{3pt}
\renewcommand{\arraystretch}{1.15}
\begin{tabular}{cccccc}
\toprule
\midrule
\multirow{2}{*}{\textbf{Scope}}
& \multirow{2}{*}{\textbf{Stage}}
& \multicolumn{2}{c}{\makecell[c]{\textbf{Gemini-3.1}\\\textbf{-Flash-Lite}}}
& \multicolumn{2}{c}{\makecell[c]{\textbf{Llama-3.3}\\\textbf{-70B}}} \\
\cmidrule(lr){3-4}
\cmidrule(lr){5-6}
& & P & R & P & R \\
\hline
\midrule
\multirow{3}{*}{Subset}
& Consensus & 54.9 & 63.1 & 66.7 & 39.9 \\
& Disagreement (before)
& 21.6 & 13.3 & 38.8 & 24.1 \\
& Disagreement (after)
& 27.8 & 8.6 & 49.2 & 19.8 \\
\midrule
\multirow{2}{*}{Overall}
& Raw candidate union
& 43.2 & 76.4 & 52.4 & 64.1 \\
& \textbf{Final prediction}
& \textbf{49.2} & \textbf{71.6}
& \textbf{59.7} & \textbf{59.8} \\
\bottomrule
\end{tabular}
\caption{Stage-wise precision (P) and recall (R)
on MIMIC-IV (\%). All recall values use the same
denominator: the total number of reference
diagnosis labels. Subset recall represents each
component's contribution to overall recall.}
\label{tab:stagewise_pr}
\vspace{-3mm}
\end{table}

\begin{table}[h]
\centering
\scriptsize
\setlength{\tabcolsep}{3.5pt}
\renewcommand{\arraystretch}{1.15}
\begin{tabular}{cccccc}
\toprule
\midrule
\multirow{2}{*}{\textbf{Backbone}}
& \multirow{2}{*}{\textbf{Transition}}
& \multicolumn{2}{c}{\textbf{Precision}}
& \multicolumn{2}{c}{\textbf{Recall}} \\
\cmidrule(lr){3-4}
\cmidrule(lr){5-6}
& & Union & Final & Union & Final \\
\hline
\midrule
\multirow{3}{*}{\makecell[c]{Gemini-3.1-Flash-Lite}}
& Persistent & 62.5 & 66.8 & 95.3 & 91.7 \\
& New        & 22.2 & 25.8 & 46.2 & 40.0 \\
& Recurrent  & 34.3 & 40.6 & 85.7 & 76.1 \\
\midrule
\multirow{3}{*}{\makecell[c]{Llama-3.3-70B}}
& Persistent & 66.0 & 68.1 & 88.1 & 85.8 \\
& New        & 26.8 & 34.3 & 26.7 & 20.0 \\
& Recurrent  & 36.2 & 44.3 & 64.3 & 52.5 \\
\bottomrule
\end{tabular}
\caption{Precision and recall before and after
adjudication by temporal transition on MIMIC-IV
(\%). Union denotes the raw candidate union;
Final denotes LONGINUS predictions.}
\label{tab:transition_pr}
\end{table}

\section{Qualitative Analysis}
\label{app:qualitative}

Tables~\ref{tab:qualitative_case_a}--\ref{tab:qualitative_case_d} show selected cases across four LLM backbones, illustrating key LONGINUS behaviors and a candidate-generation failure.

\begin{table*}[h!]
\centering
\small
\setlength{\tabcolsep}{4pt}
\renewcommand{\arraystretch}{1.15}

\begin{tabularx}{\textwidth}{@{}X@{}}
\toprule
\textbf{Patient A: MIMIC-III, Gemini-2.5-Flash-Lite;
one observed visit, predicting Visit 2.} \\
\midrule

\textbf{Patient summary.}
A male patient aged 76+ recovering from CABG,
prolonged ventilation, and dialysis-requiring renal disease
was admitted with fever, jaundice, biliary obstruction,
and pleural effusion.
ERCP was followed by pancreatitis.
His observed medical history explicitly lists peripheral
vascular disease and prostate cancer treated with
radiotherapy, although neither appeared in the observed
diagnosis-code set.
\\[4pt]

\textbf{Reference diagnoses (11).}
\textit{Persistent (7):}
Adult respiratory failure; Anemia; Biliary disease;
Cardiac dysrhythmia; Coronary atherosclerosis;
Hypertension complications; Thyroid disorders.
\textit{New (4):}
Fluid and electrolyte disorders; Peripheral atherosclerosis;
Prostate cancer; Substance-related disorders.
\\
\bottomrule
\end{tabularx}

\vspace{7pt}

\begin{minipage}[t]{0.63\textwidth}
\vspace{0pt}
\centering
\textbf{(a) Selected diagnosis predictions}

\vspace{4pt}
\scriptsize
\setlength{\tabcolsep}{3pt}
\renewcommand{\arraystretch}{1.2}
\begin{tabular*}{\linewidth}{
@{\extracolsep{\fill}}lcccc@{}
}
\toprule
\textbf{Method}
& \shortstack{\textbf{Prostate}\\\textbf{cancer}}
& \shortstack{\textbf{Peripheral}\\\textbf{atherosclerosis}}
& \shortstack{\textbf{Respiratory}\\\textbf{failure}}
& \textbf{Septicemia} \\
\midrule

\textit{Reference}
& 1 & 1 & 1 & 0 \\
\midrule

Retrieval
& \textcolor{red!70!black}{\textbf{0}}
& \textcolor{red!70!black}{\textbf{0}}
& 1
& 0 \\

Memory
& 1
& 1
& \textcolor{red!70!black}{\textbf{0}}
& \textcolor{red!70!black}{\textbf{1}} \\

\textbf{LONGINUS}
& 1
& 1
& 1
& 0 \\

Union-wide
& 1
& 1
& 1
& \textcolor{red!70!black}{\textbf{1}} \\

\bottomrule
\end{tabular*}
\end{minipage}
\hfill
\begin{minipage}[t]{0.34\textwidth}
\vspace{0pt}
\centering
\textbf{(b) Full-set performance}

\vspace{4pt}
\scriptsize
\setlength{\tabcolsep}{3pt}
\renewcommand{\arraystretch}{1.2}
\begin{tabular*}{\linewidth}{
@{\extracolsep{\fill}}lrrrr@{}
}
\toprule
\textbf{Method}
& \textbf{TP}
& \textbf{FP}
& \textbf{FN}
& \textbf{F1} \\
\midrule

Retrieval
& 7 & 6 & 4 & .583 \\

Memory
& 7 & 7 & 4 & .560 \\

\textbf{LONGINUS}
& 10 & 6 & 1 & \textbf{.741} \\

Union-wide
& 10 & 8 & 1 & .690 \\

\bottomrule
\end{tabular*}
\end{minipage}

\caption{
Illustrative case of complementary Retrieval and Memory predictions.
LONGINUS combines Memory-only New diagnoses with Retrieval-only
Persistent diagnoses while rejecting false Memory-only Septicemia.
Relative to Union-wide, it retains the same ten true positives with
fewer false positives.
Clinical history is paraphrased; red bold entries disagree with the
reference. Full-set performance includes diagnoses not shown in panel (a).
}
\label{tab:qualitative_case_a}
\end{table*}

\begin{table*}[h!]
\centering
\small
\setlength{\tabcolsep}{4pt}
\renewcommand{\arraystretch}{1.15}

\begin{tabularx}{\textwidth}{@{}X@{}}
\toprule
\textbf{Patient B: MIMIC-IV, Gemini-3.1-Flash-Lite;
two observed visits, predicting Visit 3.} \\
\midrule

\textbf{Patient summary.}
A male patient aged 41--60 with ulcerative colitis,
prior bowel perforation, and a ventral hernia underwent
staged abdominal-wall reconstruction.
Approximately 69 days after the first discharge,
he underwent proctectomy, ileal pouch construction,
diverting ileostomy, and hernia repair.
The observed notes describe recurrent
\textit{C. difficile} colitis, although Intestinal infection
was absent from the observed coding history.
\\[4pt]

\textbf{Reference diagnoses (6).}
\textit{Persistent (2):}
Hypertension; Ulcerative colitis.
\textit{Recurrent (1):}
Abdominal hernia.
\textit{New (3):}
Anemia; Cardiac dysrhythmia; Intestinal infection.
\\
\bottomrule
\end{tabularx}

\vspace{7pt}

\begin{minipage}[t]{0.63\textwidth}
\vspace{0pt}
\centering
\textbf{(a) Selected diagnosis predictions}

\vspace{4pt}
\scriptsize
\setlength{\tabcolsep}{3pt}
\renewcommand{\arraystretch}{1.2}
\begin{tabular*}{\linewidth}{
@{\extracolsep{\fill}}lcccc@{}
}
\toprule
\textbf{Method}
& \shortstack{\textbf{Ulcerative}\\\textbf{colitis}}
& \shortstack{\textbf{Abdominal}\\\textbf{hernia}}
& \textbf{Anemia}
& \shortstack{\textbf{Intestinal}\\\textbf{infection}} \\
\midrule

\textit{Reference}
& 1 & 1 & 1 & 1 \\
\midrule

Retrieval
& 1
& 1
& 1
& \textcolor{red!70!black}{\textbf{0}} \\

Memory
& 1
& 1
& \textcolor{red!70!black}{\textbf{0}}
& 1 \\

\textbf{LONGINUS}
& 1
& 1
& 1
& 1 \\

Union-wide
& 1
& 1
& \textcolor{red!70!black}{\textbf{0}}
& \textcolor{red!70!black}{\textbf{0}} \\

\bottomrule
\end{tabular*}
\end{minipage}
\hfill
\begin{minipage}[t]{0.34\textwidth}
\vspace{0pt}
\centering
\textbf{(b) Full-set performance}

\vspace{4pt}
\scriptsize
\setlength{\tabcolsep}{3pt}
\renewcommand{\arraystretch}{1.2}
\begin{tabular*}{\linewidth}{
@{\extracolsep{\fill}}lrrrr@{}
}
\toprule
\textbf{Method}
& \textbf{TP}
& \textbf{FP}
& \textbf{FN}
& \textbf{F1} \\
\midrule

Retrieval
& 4 & 3 & 2 & .615 \\

Memory
& 4 & 5 & 2 & .533 \\

\textbf{LONGINUS}
& 5 & 3 & 1 & \textbf{.714} \\

Union-wide
& 3 & 2 & 3 & .545 \\

\bottomrule
\end{tabular*}
\end{minipage}

\caption{
Illustrative case of complementary New-diagnosis prediction.
LONGINUS preserves the shared Persistent and Recurrent diagnoses
while accepting Retrieval-only Anemia and Memory-only Intestinal infection.
Union-wide removes both New diagnoses.
Clinical history is paraphrased; red bold entries disagree with the
reference. Full-set performance includes diagnoses not shown in panel (a).
}
\label{tab:qualitative_case_b}
\end{table*}

\begin{table*}[h!]
\centering
\small
\setlength{\tabcolsep}{4pt}
\renewcommand{\arraystretch}{1.15}

\begin{tabularx}{\textwidth}{@{}X@{}}
\toprule
\textbf{Patient C: MIMIC-III, GPT-OSS-120B;
two observed visits, predicting Visit 3.} \\
\midrule

\textbf{Patient summary.}
A male patient aged 61--75 with lung squamous-cell
carcinoma following right lower lobectomy,
bronchopleural fistula/empyema, invasive aspergillosis,
HIV, cardiac comorbidities, anemia, and malnutrition.
The second admission involved hypoxemic respiratory
failure and septic shock attributed to pneumonia,
requiring prolonged ventilation, tracheostomy,
and gastrostomy.
\\[4pt]

\textbf{Reference diagnoses (12).}
\textit{Persistent (10):}
Adult respiratory failure; Congestive heart failure,
non-hypertensive; Coronary atherosclerosis;
Fluid and electrolyte disorders; HIV infection;
Bronchus/lung cancer; Mycoses; Septicemia; Shock; Pleurisy.
\textit{Recurrent (1):} Anemia.
\textit{New (1):} Aspiration pneumonia.
\\
\bottomrule
\end{tabularx}

\vspace{7pt}

\begin{minipage}[t]{0.63\textwidth}
\vspace{0pt}
\centering
\textbf{(a) Selected diagnosis predictions}

\vspace{4pt}
\scriptsize
\setlength{\tabcolsep}{3pt}
\renewcommand{\arraystretch}{1.2}
\begin{tabular*}{\linewidth}{
@{\extracolsep{\fill}}lcccc@{}
}
\toprule
\textbf{Method}
& \textbf{Septicemia}
& \textbf{Pleurisy}
& \shortstack{\textbf{Pulmonary}\\\textbf{heart disease}}
& \shortstack{\textbf{Aspiration}\\\textbf{pneumonia}} \\
\midrule

\textit{Reference}
& 1 & 1 & 0 & 1 \\
\midrule

Retrieval
& 1
& 1
& \textcolor{red!70!black}{\textbf{1}}
& \textcolor{red!70!black}{\textbf{0}} \\

Memory
& 1
& \textcolor{red!70!black}{\textbf{0}}
& 0
& \textcolor{red!70!black}{\textbf{0}} \\

\textbf{LONGINUS}
& 1
& 1
& 0
& \textcolor{red!70!black}{\textbf{0}} \\

Union-wide
& \textcolor{red!70!black}{\textbf{0}}
& 1
& \textcolor{red!70!black}{\textbf{1}}
& \textcolor{red!70!black}{\textbf{0}} \\

\bottomrule
\end{tabular*}
\end{minipage}
\hfill
\begin{minipage}[t]{0.34\textwidth}
\vspace{0pt}
\centering
\textbf{(b) Full-set performance}

\vspace{4pt}
\scriptsize
\setlength{\tabcolsep}{3pt}
\renewcommand{\arraystretch}{1.2}
\begin{tabular*}{\linewidth}{
@{\extracolsep{\fill}}lrrrr@{}
}
\toprule
\textbf{Method}
& \textbf{TP}
& \textbf{FP}
& \textbf{FN}
& \textbf{F1} \\
\midrule

Retrieval
& 11 & 8 & 1 & .710 \\

Memory
& 10 & 5 & 2 & .741 \\

\textbf{LONGINUS}
& 11 & 5 & 1 & \textbf{.786} \\

Union-wide
& 9 & 7 & 3 & .643 \\

\bottomrule
\end{tabular*}
\end{minipage}

\caption{
Illustrative case of consensus preservation and disagreement adjudication.
LONGINUS preserves consensus Septicemia, accepts Retrieval-only Pleurisy,
and rejects false Retrieval-only Pulmonary heart disease.
Neither view proposes the target Aspiration pneumonia, illustrating a
candidate-generation failure.
Clinical history is paraphrased; red bold entries disagree with the
reference. Full-set performance includes diagnoses not shown in panel (a).
}
\label{tab:qualitative_case_c}
\end{table*}


\begin{table*}[t]
\centering
\small
\setlength{\tabcolsep}{4pt}
\renewcommand{\arraystretch}{1.15}

\begin{tabularx}{\textwidth}{@{}X@{}}
\toprule
\textbf{Patient D: MIMIC-IV, Llama-3.3-70B;
four observed visits, predicting Visit 5.} \\
\midrule

\textbf{Patient summary.}
A male patient aged 76+ with insulin-treated diabetes,
severe peripheral vascular disease, coronary artery disease
after CABG, and hypothyroidism had repeated vascular
admissions.
Earlier visits involved ischemic foot disease, infection,
revascularization, bypass-graft infection and removal,
non-healing ulcers, and bypass revision.
The fourth observed admission concerned left-leg ischemia
and planned bypass surgery.
Gangrene was coded in Visit 3 but not Visit 4.
\\[4pt]

\textbf{Reference diagnoses (6).}
\textit{Persistent (5):}
Conduction disorders; Coronary atherosclerosis;
Diabetes mellitus without complications; Skin ulcer;
Thyroid disorders.
\textit{Recurrent (1):} Gangrene.
\textit{New (0).}
\\[4pt]

\textbf{Temporal and tail signal.}
Gangrene coding across observed Visits 1--4:
0, 0, 1, 0; target: 1.
Gangrene is a Tail label.
These states denote diagnosis coding, not necessarily
clinical resolution or reactivation.
\\
\bottomrule
\end{tabularx}

\vspace{7pt}

\begin{minipage}[t]{0.63\textwidth}
\vspace{0pt}
\centering
\textbf{(a) Selected diagnosis predictions}

\vspace{4pt}
\scriptsize
\setlength{\tabcolsep}{3pt}
\renewcommand{\arraystretch}{1.2}
\begin{tabular*}{\linewidth}{
@{\extracolsep{\fill}}lcccc@{}
}
\toprule
\textbf{Method}
& \shortstack{\textbf{Skin}\\\textbf{ulcer}}
& \textbf{Gangrene}
& \shortstack{\textbf{Cardiac}\\\textbf{dysrhythmia}}
& \textbf{Hyperlipidemia} \\
\midrule

\textit{Reference}
& 1 & 1 & 0 & 0 \\
\midrule

Retrieval
& 1
& \textcolor{red!70!black}{\textbf{0}}
& 0
& \textcolor{red!70!black}{\textbf{1}} \\

Memory
& \textcolor{red!70!black}{\textbf{0}}
& 1
& \textcolor{red!70!black}{\textbf{1}}
& \textcolor{red!70!black}{\textbf{1}} \\

\textbf{LONGINUS}
& 1
& 1
& 0
& \textcolor{red!70!black}{\textbf{1}} \\

Union-wide
& 1
& 1
& \textcolor{red!70!black}{\textbf{1}}
& \textcolor{red!70!black}{\textbf{1}} \\

\bottomrule
\end{tabular*}
\end{minipage}
\hfill
\begin{minipage}[t]{0.34\textwidth}
\vspace{0pt}
\centering
\textbf{(b) Full-set performance}

\vspace{4pt}
\scriptsize
\setlength{\tabcolsep}{3pt}
\renewcommand{\arraystretch}{1.2}
\begin{tabular*}{\linewidth}{
@{\extracolsep{\fill}}lrrrr@{}
}
\toprule
\textbf{Method}
& \textbf{TP}
& \textbf{FP}
& \textbf{FN}
& \textbf{F1} \\
\midrule

Retrieval
& 4 & 4 & 2 & .571 \\

Memory
& 2 & 5 & 4 & .308 \\

\textbf{LONGINUS}
& 6 & 4 & 0 & \textbf{.750} \\

Union-wide
& 6 & 5 & 0 & .706 \\

\bottomrule
\end{tabular*}
\end{minipage}

\caption{
Illustrative long-history case involving a recurrent Tail diagnosis.
LONGINUS combines Retrieval-only Persistent diagnoses with Memory-only
recurrent Gangrene and rejects false Memory-only Cardiac dysrhythmia.
Both LONGINUS and Union-wide recover all reference diagnoses, but
LONGINUS retains one fewer false positive.
Clinical history is paraphrased; red bold entries disagree with the
reference. Full-set performance includes diagnoses not shown in panel (a).
}
\label{tab:qualitative_case_d}
\end{table*}

\end{document}